\documentclass[journal]{vgtc}                

\vgtcinsertpkg

\ifpdf
  \pdfoutput=1\relax                   
  \usepackage{graphicx}                
  \DeclareGraphicsExtensions{.pdf,.png,.jpg,.jpeg} 
\else
  \usepackage{graphicx}                
  \DeclareGraphicsExtensions{.eps}     
\fi%

\graphicspath{{figures/}{pictures/}{images/}{./}} 

\usepackage{microtype}                 
\PassOptionsToPackage{warn}{textcomp}  
\usepackage{textcomp}                  
\usepackage{mathptmx}                  
\usepackage{times}                     
\usepackage{cite}                      
\usepackage{tabu}                      
\usepackage{booktabs}                  

\usepackage{xspace}

\usepackage{paralist}
\usepackage{enumitem}
\usepackage{gensymb}
\usepackage{amsmath}
\usepackage{amssymb}

\usepackage{moresize}

\usepackage{wrapfig}

\usepackage{scalerel}

\newcommand{\toolname}{{\texttt{PSEUDo}}\xspace}
\newcommand{\toolnames}{{\texttt{PSEUDo's}}\xspace}

\newcommand{\R}[1]{{\textbf{R#1}}\xspace}
\newcommand{\codeurl}{{\url{https://git.science.uu.nl/vig/sublinear-algorithms-for-va/locality-sensitive-hashing-visual-analytics}}\xspace}

\definecolor{mred}{rgb}{.80,.12,.30}
\definecolor{MRED}{rgb}{.80,.12,.30}
\definecolor{grey}{rgb}{0.5,0.5,0.5}
\definecolor{lgrey}{rgb}{0.7,0.7,0.7}
\definecolor{purple}{rgb}{.75,0,.85}
\definecolor{pistachio}{rgb}{0.58, 0.77, 0.45}
\definecolor{traingreen}{HTML}{4CAF50}
\definecolor{myorange}{rgb}{0.94, 0.36, 0.13}

\newif\ifnotes
\notestrue

\let\origcite\cite

\renewcommand{\cite}[1]{\ifnotes\mbox{\origcite{#1}}\else \origcite{#1}\fi}

\usepackage[nolist]{acronym}

\usepackage{hyperref}
\usepackage[capitalise,nameinlink]{cleveref}
\usepackage{subcaption}

\usepackage{float}

\usepackage{ragged2e}

\usepackage{adjustbox}

\onlineid{0}

\vgtccategory{Research}
\vgtcpapertype{please specify}

\title{\toolname: Interactive Pattern Search in Multivariate Time Series with Locality-Sensitive Hashing and Relevance Feedback}

\author{Yuncong Yu, Dylan Kruyff, Tim Becker, Michael Behrisch}
\authorfooter{
\item
 Yuncong Yu is with IAV GmbH. E-mail: yuncong.yu@iav.de.
\item
 Dylan Kruyff is with Utrecht University.
 E-mail: d.l.w.kruyff@students.uu.nl 
\item
 Tim Becker is with IAV GmbH. E-mail: tim.becker@iav.de.
\item
 Michael Behrisch is with Utrecht University.
 E-mail: m.behrisch@uu.nl.
}

\shortauthortitle{Yu \MakeLowercase{\textit{et al.}}: \toolname: Multivariate Time Series Exploration Using Locality Sensitive Hashing}

\abstract{
We present \toolname, an adaptive feature learning technique for exploring visual patterns in multi-track sequential data. Our approach is designed with the primary focus to overcome the uneconomic retraining requirements and inflexible representation learning in current deep learning-based systems. 
Multi-track time series data are generated on an unprecedented scale due to increased sensors and data storage. 
These datasets hold valuable patterns, like in neuromarketing, where researchers try to link patterns in multivariate sequential data from physiological sensors to the purchase behavior of products and services. But a lack of ground truth and high variance make automatic pattern detection unreliable.
Our advancements are based on a novel query-aware Locality-Sensitive Hashing technique to create a feature-based representation of multivariate time series windows. Most importantly, our algorithm features sub-linear training and inference time. We can even accomplish both the modeling \emph{and} comparison of 10,000 different 64-track time series, each with 100 time steps (a typical EEG dataset) under 0.8 seconds. This performance gain allows for a rapid relevance feedback-driven adaption of the underlying pattern similarity model and enables the user to modify the speed-vs-accuracy trade-off gradually.
We demonstrate \toolname's superiority in terms of efficiency, accuracy, and steerability through a quantitative performance comparison and a qualitative visual quality comparison to the state-of-the-art algorithms in the field. Moreover, we showcase \toolname's usability through a case study demonstrating our visual pattern retrieval concepts in a large meteorological dataset. 
We find that our adaptive models can accurately capture the user's notion of similarity and allow for an understandable exploratory visual pattern retrieval in large multivariate time series datasets.

} 


\CCScatlist{ 
 \CCScat{K.6.1}{Management of Computing and Information Systems}%
{Project and People Management}{Life Cycle};
 \CCScat{K.7.m}{The Computing Profession}{Miscellaneous}{Ethics}
}

\teaser{
  \centering
  \includegraphics[width=\linewidth]{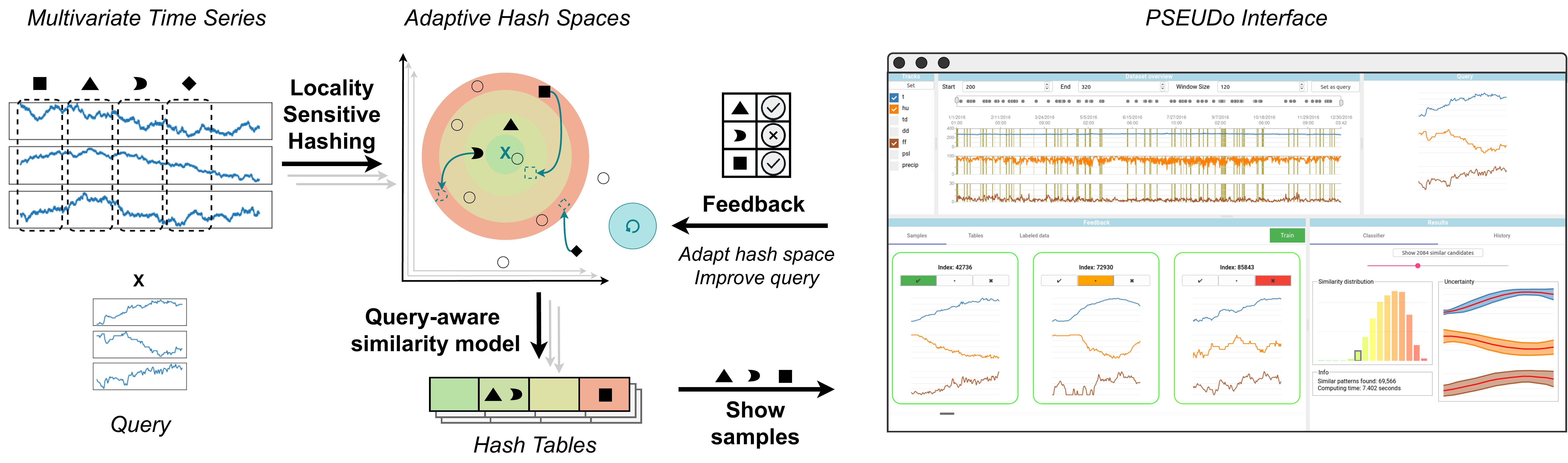}
  \caption{\textbf{The \toolname system.} Based on an adaptive Locality-Sensitive Hashing algorithm (can be typically learned in near real-time), we extract a visual pattern similarity model adopted for interactive exploration. Using a relevance-feedback mechanism, the user can interactively modify the current notion of pattern interestingness 
  or give more impact on the accuracy-vs-speed trade-off in the model's retrieval uncertainty.
  }
  \label{fig:teaser}
}

\begin{document}

\begin{acronym}
    \acro{dba}[DBA]{Dynamic Time Warping Barycenter Averaging}
    \acro{dtw}[DTW]{Dynamic Time Warping}
    \acro{dtwd}[DTW\textsubscript{D}]{Dynamic Time Warping Dependent}
    \acro{ed}[ED]{Euclidean Distance}
    \acro{lsh}[LSH]{Locality-Sensitive Hashing}
    \acro{mts}[MTS]{Multivariate Time Series}
    \acro{paa}[PAA]{Piecewise Aggregate Approximation}
    \acro{sax}[SAX]{Symbolic Aggregate approXimation}
    \acro{qalsh}[QALSH]{Query-Aware Locality Sensitive Hashing}
    \acro{uts}[UTS]{Univariate Time Series}
\end{acronym}
\firstsection{Introduction}
\label{sec:introduction}
\maketitle

Although Deep Learning models have led to a rejuvenation of artificial intelligence along with tremendous progress on practical problems, the non-scalable training process inhibits their application in highly exploratory and dynamic use cases. 
Consequently, these advanced techniques are limited to offline preprocessing machineries, specifically tailored to a few data analysis tasks \cite{DBLP:journals/cgf/ChatzimparmpasM20, DBLP:journals/cgf/EndertRTWNBR17}. 
In the context of exploratory analysis, however, we \emph{need} more flexible solutions, as the nuanced interplay between exploration and exploitation requires the ability to constantly (re-)define our exploration models\cite{DBLP:journals/tvcg/SachaSSKEK14}. 
In this line of research, and particularly in this paper, we are therefore 
choosing a complementary path and explore sub-linear machine learning algorithms whose training times 
are a fraction of the time needed for training deep neural networks. 
This speedup 
opens up new opportunities to promptly react to the newly occurring problem sets by, e.g., constant model (re-)evaluation/training or on-the-fly relevance feedback adaption of the underlying exploration model.

With \toolname, we present a system for interactive visual pattern exploration in large \ac{mts}. 
We chose this domain since it exposes a range of interesting challenges from the Visual Analytics and the technical perspective.
From the Visual Analytics point of view, pattern search in \ac{mts} data is inherently a user-centered problem as the data analysts' domain understanding is the key factor in the pattern exploration and interpretation process\cite{DBLP:series/hci/AignerMST11}. A study by Correll and Gleicher\cite{M.Correll.2016b} shows that no single algorithm accounts for human judgments of time series similarity. This dilemma, coined \textbf{biased similarity}, implies that a canonical representation (featurization) of patterns, or what a retrieval index can capture, is only part of the solution. 
The remainder is retrieving interesting patterns based on the user's \emph{notion of similarity}. 
To define similarity, researchers have created various distance metrics\cite{DBLP:conf/cbmi/GregorLSSB15} and, similar to our approach in \toolname, (adaptive) search and manifold learning strategies\cite{DBLP:conf/ieeevast/BrownLBC12, dennig2019fdive, DBLP:journals/tvcg/MaTWGPC20}. 

The next challenge for pattern search in \ac{mts} is \textbf{cognitive biases}, particularly the
\href{https://en.wikipedia.org/wiki/Confirmation_bias}{confirmation bias}, i.e., people tend to search for, interpret, favor, and recall information in a way that confirms their prior beliefs, 
and more generally,  
\href{https://en.wikipedia.org/wiki/Pareidolia}{Pareidolia}, people's tendency for an incorrect perception of a stimulus as a pattern\cite{DBLP:journals/tvcg/WickhamCHB10, DBLP:journals/tvcg/DimaraFPBD20}.
Even experienced domain experts oversee striking pattern(-combinations) only because they are biased or overwhelmed by information overload.
In \toolname, we can circumvent these problems by combining interactive pattern detection with unbiased or even bias-counteracting recommendation algorithms.
As we will present in this work, this approach has several advantages. For example, since the \textit{interestingness of patterns} is not solely determined by the computer, we can offer a mixed-initiative system in which the user has full governance over the pattern retrieval process.
Another advantage is the exploration models' unbiased representation of the feature or latent space capturing the dataset's information.
As such, we can always recommend unexplored pattern spaces, hence actively combating the expert's cognitive biases.

As a technical challenge, it is evident that the sheer \textbf{size of \ac{mts} datasets} requires advanced search and retrieval algorithms to capture the potentially inter-linked multi-track patterns; i.e., a 60min EEG epilepsy/seizure screening dataset captured on 32-channels with 1000Hz amounts to  $\sim$3.7GB of raw data; typically, hundreds of subjects participate in these studies\cite{jeong2004eeg}. Even after applying sophisticated data compression techniques and efficient data structures, pattern search still requires an enormous number of comparisons, 
where the number is the product of time steps and dimension of measured features. 
A human alone is unable to accomplish this task. 
However, even for computers, this investigation is not trivial, as demonstrated by a plethora of approaches targeting the automated analysis of \ac{uts} \cite{hamilton2020time, Lekschas.2020, fiterau2016similarity, DBLP:conf/bigcom/GuoWJ17, LimTSForcasting2020}. The investigation of \ac{mts}, as in this paper, is inevitably more complex. 
Even though \ac{mts} are ``just'' an agglomeration of multiple tracks or dimensions, and each track itself is just an agglomeration of several data points (feature expressions), we cannot treat these points independently. 
All values are potentially 
correlated and form an intertwined \textbf{dependency network of temporal patterns}, one example being the (time-delayed) gene-protein reaction chains 
studied in computational ecogenomics\cite{thessen2020predicting}. 
This property distinguishes \ac{mts} data from most other data types and requires custom solutions to general data analysis problems.

We propose \toolname, a scalable (visual) pattern analysis technique based on the conceptual idea of relevance feedback-driven exploration\cite{M.Behrisch.2014},
allowing us to tackle all four challenges.
Our method focuses on a human-in-the-loop approach in which we combine a query-aware adaption of \ac{lsh} with a novel feedback-driven active learning algorithm. 
Our feature learning and indexing technique allows for capturing and adapting the notion of similarity between visual patterns in \ac{mts} in near real-time. 
\ac{lsh} refers to a set of probabilistic functions, known as \acs{lsh} families, to hash data points into buckets.  
Similar data points are located in the same buckets (with high probability), while dissimilar data is likely to be placed in different buckets\cite{DBLP:conf/iclr/KitaevKL20}.
If \ac{lsh} functions are defined for a metric space, like in our case, the buckets even imply a distance between each other\cite{gionis1999similarity}.
\toolname actively makes use of these ideas and maps time series windows into buckets in hash tables. Hash collisions indicate groups of similar time series window candidates. 
The final result of \ac{lsh} is obtained by an ensemble of all hash tables\cite{Indyk.1998b}.
In \toolname, we exploit this ensemble step to convey the subjectively perceived similarity between visual patterns to our \acs{lsh} model. 
We implemented this relevance feedback mechanism through feature weighting, attaching more importance to positively labeled time series samples and hash tables that better reflect our notion of similarity. 
We make \toolname accessible through a prototypical interface and evaluate the potential of our idea in a series of technical and real-world evaluation scenarios.
Our logical first step is to prove our \ac{lsh} algorithm's capabilities to capture visual features in sequential datasets. 
To ensure comparability, we conducted six quantitative experiments using 
standard information retrieval metrics, like recall and accuracy, on four \ac{mts} datasets and compare ourselves with four state-of-the-art TS analysis algorithms, \ac{dtwd}\cite{MohammadShokoohiYekta.}, \ac{sax}\cite{Lin.2003}, \ac{ed}. 
We find that our model retrieves patterns with similar accuracy but significantly outperforms all other approaches in terms of processing and retrieval speed whenever the track-size increases or the pattern shape matching process is getting non-trivial. 
Additionally, we conducted one \ac{uts} experiment with the aforementioned algorithms and the recently introduced autoencoder-based technique PEAX\cite{Lekschas.2020}. 
We achieve over \textbf{2160x faster processing speed} with our slowest but most accurate implementation variant compared to PEAX, thus enabling interactive and steerable retrieval of visual patterns now also in large and multi-track temporal datasets. 
We evaluate \toolname's usability in a series of case studies, which we will demonstrate in the paper
and the paper appendix. 
These case studies demonstrate how analysts can use \toolname to capture and adapt their subjectively perceived pattern similarity effectively, how this adaption blends into their analysis, and how it contributes to a faster and user-centered exploration.

\section{Requirement Analysis and Background}
\label{sec:background}

This section gives an overview of the domain, its challenges, and recent advancements. To keep the didactic flow, 
we decided to layout \toolname's primary embedding 
into the relevant academic topics 
in this section and elaborate on the gap in the related work in \autoref{sec:relatedwork}.

\subsection{Requirements for Multi-Track Time Series Exploration}
\label{subsec:requirements_for_multi_track_time_series_exploration}
\toolname has been developed for the analysis of \ac{mts} datasets. 
These datasets find wide application in various disciplines, such as marketing and finance, biological and medical research, network security, IoT, or video and audio analysis\cite{DBLP:journals/jmlr/KhaleghiRMP16, DBLP:journals/ijon/AhmadLPA17}. 
We conducted a broad, serendipitous literature research spanning various domains starting from the past years IEEE VIS and Eurographics EuroVIS papers. We collected an initial list of requirements for \ac{mts} data visualizations and analysis. 
To be of practical use for the broad public, we generalize our criteria to be domain-independent based on our experience in the field. 

\begin{enumerate}\setlength\itemsep{0.05em}
    \item[\textbf{R1}] \textbf{Scalable time complexity}. The query-relevant size of \ac{mts} grows exponentially with the number of time steps and tracks. Processing algorithms should, therefore, scale (sub-)linearly with respect to the query size, dataset size, and number of tracks.
    \item[\textbf{R2}] \textbf{Adaptive similarity definition.} Similarity perception differs from user to user\cite{DBLP:conf/chi/PandeyKFBB16}. As a result, the similarity measure should be adjustable and should converge to the user's wishes.
    \item[\textbf{R3}] \textbf{Accurate pattern retrieval.} Based on a notion of similarity, the pattern retrieval algorithm should find the majority of similar patterns with high accuracy, e.g., $>90\%$.
    On top of that, the algorithm should accurately approximate the degree of similarity between a pattern and the query.
    \item[\textbf{R4}] \textbf{Interactive querying and model steering.} On the one hand, users need a mechanism to define a visual search pattern, for instance, through query-by-sketch or query-by-example\cite{M.Correll.2016b}, and on the other hand,  a way to steer the retrieval process. One opportunity here is active learning-based approaches that rely on the user's explicit relevance feedback, such as in\cite{M.Behrisch.2014, dennig2019fdive, Lekschas.2020}. 
    \item[\textbf{R5}] \textbf{Understandable process.} To understand the data and algorithm changes, transparent and user-friendly visualization of the model's internal mechanism \emph{and} results is necessary.
    \item[\textbf{R6}] \textbf{Free exploration.} The user should be able to engage freely with the data to find and explore patterns. As user feedback may change the direction of the pattern search, users need a way to revert or adapt query- and similarity definitions flexibly.
\end{enumerate}

Our requirement analysis lays the foundation of \toolname's design and will be referred to throughout this work. 
We distinguish two categories: \textit{technical/algorithmic} and \textit{Visual Analytics}. 
On the technical side, we highlight the need for a scalable \textbf{R1}, accurate \textbf{R3}, steerable \textbf{R4} (through relevance feedback) and understandable \textbf{R5} visual \ac{mts} analysis processes. From the Visual Analytics perspective, we require that our interface supports and fosters querying \textbf{R4}, feedback interaction \textbf{R2}, exploration \textbf{R6}, and transparency \textbf{R5}. 

Our requirement gathering approach led us to an interesting secondary finding that we would like to share: Although several domains have already put a great accent on time series pattern visualization and mining\cite{DBLP:series/hci/AignerMST11, buono2005interactive, motif, ding2008querying}, a thorough and systematic \emph{\ac{mts} task taxonomy} is still missing. We plan to fill this gap in the literature soon. 

\subsection{\toolnames Powerhouse: \acl{lsh}}
\label{subsec:pseudos_powerhouse}


The bedrock of our computational approach is laid on \ac{lsh}\cite{Indyk.1998b}.
In general, hashing-based algorithms excel in low time-complexity during model building and inference and a small model footprint
, especially for large datasets. 
\ac{lsh} inherits the high speed while differing from the other hashing algorithms in that it maps similar data objects to close hash codes.
This enables its application in a wide range of data mining problems, like nearest neighbor search\cite{Indyk.1998b, Huang.2015}, hierarchical clustering\cite{Koga.2007, Cochez.2015} and near-duplicate detection\cite{Das.2007}.

The classical (vector-based) \ac{lsh} conceives data objects in the database as points in a high dimensional space. 
The hash function describes a set of random hyperplanes cutting the space into disjoint subspaces\cite{Indyk.1998b}.
Each subspace corresponds to a bucket in the hash table,
meaning that the entry points in the same subspace fall into the same hash table bucket.
To reduce false positives and false negatives (especially the latter), multiple hash functions, i.e., multiple sets of hyperplanes, are used.
All the entries that collide with the query under at least one hash function are treated as candidates.
Subsequently, the candidates can be filtered in a second stage to suppress false positives.

Since \ac{lsh} is considerably fast, one can even postpone part of the model building process to the querying phase,
as suggested by \ac{qalsh} in the context of high-dimensional data retrieval\cite{Huang.2015}.
By using the query's hash code as an anchor for partitioning the hash tables, \ac{qalsh} yields better accuracy while accelerating the hashing with simpler hash functions.


Recently, \ac{lsh} received attention also for time series indexing. Up to 20x faster processing speeds with nearly the same accuracy as state-of-the-art algorithms could be achieved\cite{ChenLuo.2017}.
Shortly after, a query-aware \ac{lsh} approach for multivariate cases was demonstrated by Yu et al.\cite{ChenyunYu.2019}. This work builds the cornerstones of our system. 
In \toolname, we contribute to the \ac{qalsh}-\ac{mts} algorithm of Yu et al. with a structural modification:
Our method uses a distance metric on the hash functions instead of the original data, thus turning the approach into an approximate retrieval algorithm (we will detail this aspect in \autoref{subsec:modelmts}). This modification has the benefit of introducing a relevance feedback mechanism to tackle the problem of subjectivity in the notion of similarity, i.e., no single similarity measure accounts for human judgment of time series similarity\cite{M.Correll.2016b}. Moreover, our processing time stays constant as the number of tracks increases (in the original \ac{qalsh}-\ac{mts} algorithm we have a linear growth). On the downside, our approximations naturally lead to a slight decrease in accuracy, which is, on the other hand, only an academic consideration in our exploratory data analysis setting. 

\subsection{Recent Developments: Deep Learning-based Interactive Visual Pattern Search}
One development has sparked particularly our critical reflection with this analysis field and specifically the of  topic economic efficiency. 
Recently, novel deep-learning-based interactive visual pattern search algorithms were introduced; see, e.g., \cite{DBLP:journals/tvcg/SpinnerSSE20, DBLP:journals/tvcg/HohmanKPC19} for an overview. 
One prominent and thorough example is PEAX, ``a novel feature-based technique for interactive visual pattern search in sequential data''\cite{Lekschas.2020} by Lekschas et al. PEAX uses an autoencoder to extract the TS representations for each window (feature vector) 
from a dataset. 
After this (static) featurization stage, an adaptive classifier receives binary relevance feedback from a user. 
It uses this feedback to influence the pattern retrieval algorithm by adapting the importance of latent-space dimensions. 
Similar approaches were presented, e.g., for image data\cite{DBLP:journals/tvcg/WangTSPDHKC21}, TS data\cite{DBLP:conf/bigcom/GuoWJ17, fiterau2016similarity, RobertsMusic2017}, or even information visualizations\cite{DBLP:journals/tvcg/MaTWGPC20, DBLP:journals/cga/Haleem0PWQ19}.

We claim that Deep Learning-, and particularly autoencoder-based concepts are only sustainable for VA in pattern-wise strongly confined application fields.
These are fields in which visual pattern occurrences do not or only linearly influence each other, such as in univariate or single-track time series. 
In \ac{mts} data, patterns affect each other through delay or spread and even over multiple tracks.
One possible solution for autoencoders, in this case, is to concatenate individual channels and featurize them as one. 
But not only does this invoke the curse of dimensionality, but it also eliminates the possibility of capturing inter-channel correlations.
This problem is general and exists in several other domains, such as textual/discourse, network, or video data analysis.

Secondly, the use of autoencoders imposes further drawbacks. 
An autoencoder training can take hours and even up to days to finish\cite{Lekschas.2020} and is then fixed to a specific window or kernel size. 
This is vastly inconvenient in terms of exploration flexibility. 
Every time the user wants to change, e.g., the pattern's window size, the autoencoder needs to be retrained. 
Furthermore, the autoencoder is trained on a \textit{specific} domain/dataset. 
For MTS, this means that the model might be able to extract essential features from one channel but might perform poorly on another track. 

We explore an orthogonal approach to this problem based on significantly more light-weight hashing algorithms in this work.
Note that we discuss further related work in \autoref{sec:relatedwork} at the end of this paper.
\section{Classifier Learning using Steerable Locality-Sensitive Hashing based on Relevance Feedback}
\label{sec:classifierlearning}

This section introduces our two primary technical contributions: (1) a novel feedback-driven classifier learning that is based on (2) a novel \ac{qalsh} adaption.
Not only does this method allow pattern search in large \ac{mts} to be fast enough for real-time human interaction, 
but our concept of a steerable \ac{lsh} is well transferable to other applications.
Please note that unless otherwise specified, we use \ac{lsh} as an abbreviation for \ac{qalsh} in subsequent sections. 

\begin{figure}[H]
    \centering
    \includegraphics[width=\linewidth]{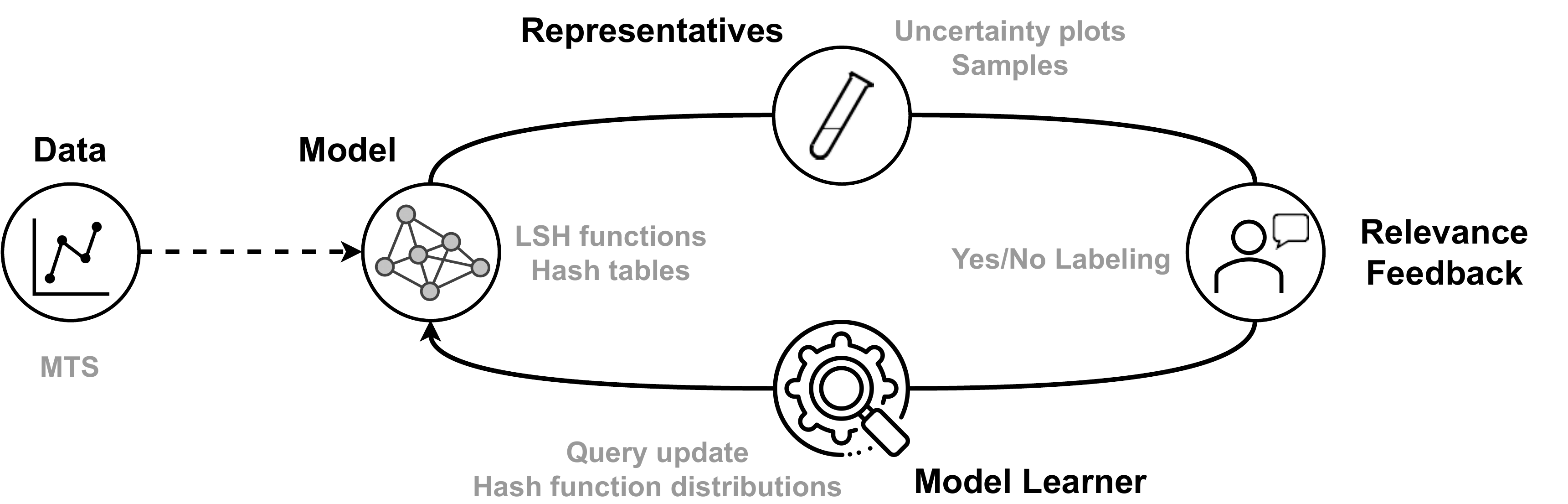}
    \caption[Conceptual overview]{
    The conceptual model of our proposed method is 
    (1) an application of FDive\cite{M.Behrisch.2014} for \ac{mts} retrieval,  
    (2) it models data with \ac{lsh}, 
    (3) sample the outcomes and 
    (4) invite the user to review the model's similarity understanding, 
    (5) it optimizes the \ac{lsh} functions based on the user feedback.
    }
    \label{fig:conceptual_model}
\end{figure}

We derive our conceptual model, depicted in \autoref{fig:conceptual_model}, from FDive, the feedback-driven preference model learning approach proposed by Behrisch et al.\cite{M.Behrisch.2014}. 
In light of this conceptual model, we assume that the randomly initialized parameters in \ac{lsh} functions are trainable,
creating room for coping with the ambiguity of time series similarity.

\begin{figure*}[ht]
    \centering
    \includegraphics[width=0.85\linewidth]{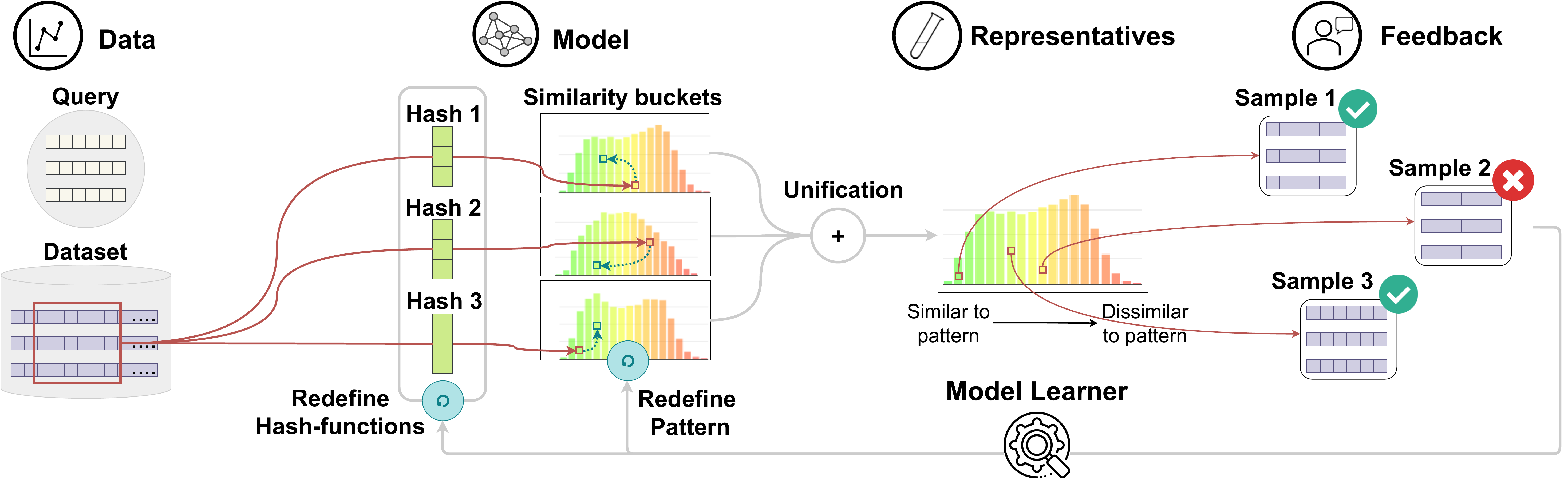}
    \caption[Pipeline]{
    \toolname \textbf{Processing Pipeline}:
    (1) We receive a query and preprocessed time series windows as our inputs, 
    (2) hash them into hash table buckets representing the similarity distribution to the query (green depicts similar, red suggests dissimilarity), 
    (3) draw representatives from both similar and dissimilar buckets, 
    (4) receive user feedback, 
    (5) and update the hash functions and the query pattern specificity accordingly. 
    }
    \label{fig:pipeline}
\end{figure*}

In our conceptual model, input \ac{mts} data is modeled with \ac{lsh} functions and stored in query-aware hash tables.
To guarantee a sensible data modeling on top of the probabilistic nature of \ac{lsh} functions (\R{1}),
and combat subjectiveness of similarity (\R{2}), we draw a sample from the hashed time series. 
This sample set gets checked and (re-)evaluated by the user on the "average" shape and variance per bucket.
This relevance feedback is actively contributing to the interactive pattern search in that \toolname learns to estimate the user's similarity notion 
and interpret it as feature/track importance. 
Due to the size and complexity of \ac{mts} data, such an open and adaptable exploration process (\R{4}, \R{6}) was unthinkable before within our self-imposed scalability and performance limits.
Moreover, we can see that such a user-in-the-loop (active) learning approach not only improves the overall retrieval performance (\R{3}) but also significantly adds to the user understandability (\R{5})
of finding and exploring patterns in \ac{mts}.

To ensure a precise and concise description of our work, we introduce the following notations: 
We use lower case letters for scalars, e.g., $a$; subscript $i$ for time steps and $ j $ for track indices, e.g., $a_i$ and $ r $ for iterations of training, e.g., $w_r$; upper case letter are used for matrices, e.g., $A$; 
Vectors are denoted with arrows, like $ \vec{a} $. 
To describe element-wise operations, vectors and matrices can also be represented per their entries like $ \{ a_j \} $. We use $\mathbb{C}$ to denote sampling candidates, with $\mathbb{C}^{+/-}$ being positively or negatively labeled items.
Next, we will explain how we transform our conceptual model into the concrete \toolname pipeline depicted in \autoref{fig:pipeline}. 

\subsection{Modeling \acl{mts}}
\label{subsec:modelmts}

\begin{figure}[H]
    \centering
    \includegraphics[width=\linewidth]{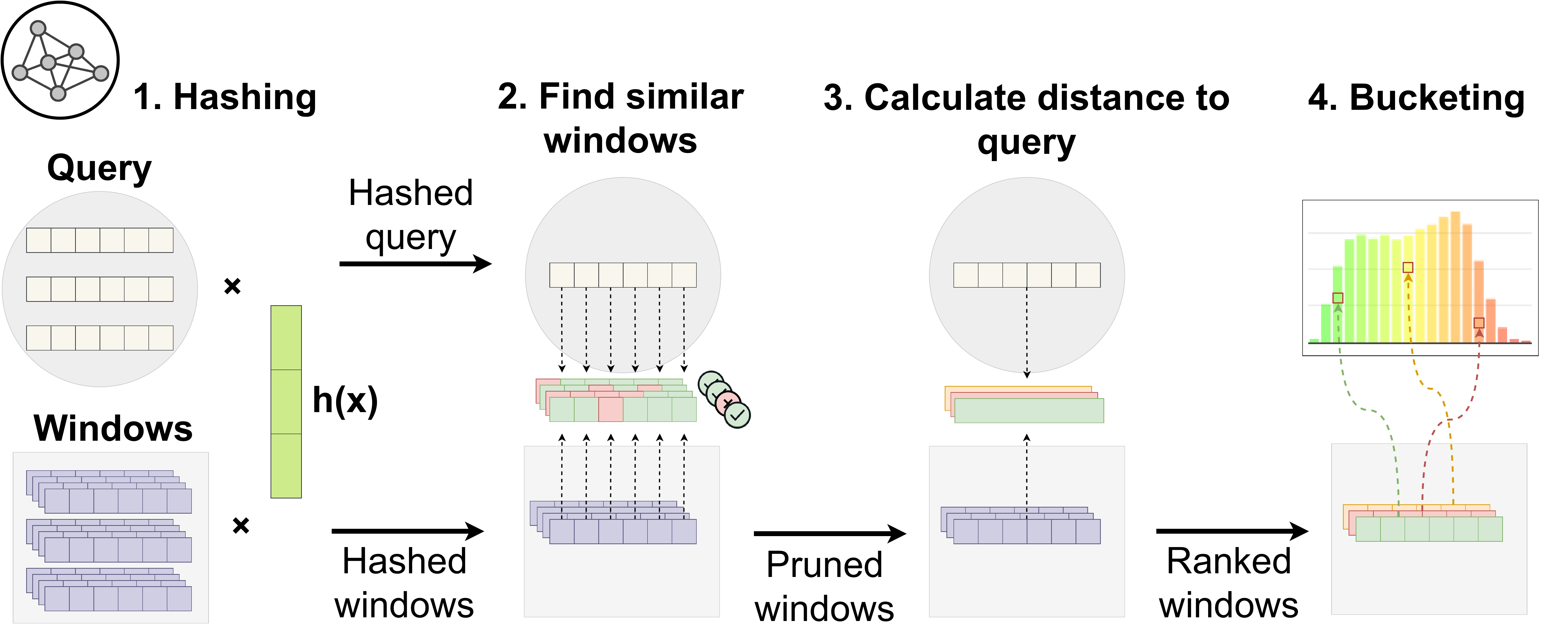}
    \caption[Modeling process]{
    \textbf{Modeling process (detail)}: 
    (1) We hash all windows including the query to a single \ac{uts},
    (2) prune dissimilar windows with the algorithm in\cite{ChenyunYu.2019}, 
    (3) rank the remaining windows based on their \ac{dtw} or \ac{ed} similarity to the query, 
    and (4) partition the ranked windows into hash tables buckets.
    }
    \label{fig:model}
    \vspace{-0.8em}
\end{figure}

As \autoref{fig:model} depicts, the modeling process begins with partitioning the $ d $-dimensional \ac{mts} $ S = \{{\vec{s}_i} \in \mathbb{R}^d\} $, $ i = 1,2,...,n $ of length $ n $, in which to search for the \textbf{Query} $ Q = \{\vec{q}_i \in \mathbb{R}^d\} $, $ i = 1,2,...,t $, with a sliding \textbf{Window} of size $ t $ (given by $ Q $).
This approach converts the local pattern search problem into a more frequently addressed problem, namely 
time series indexing\cite{faloutsos1994fast,AndreJonsson.2002,mtspatternsearch}.
Subsequently, the windows and the query are normalized to focus on the pattern shape 
and simplify hashing.

Generally, the modeling follows the approach proposed by Yu et al. in\cite{ChenyunYu.2019}, 
where $ l $ compound \textbf{hash functions h(x)} are generated, each corresponding to a hash table or a classifier
(in \autoref{fig:model} omitted for readability). 
We use ``\textit{compound hash function}'', ``\textit{hash table}", and ``\textit{classifier}" as synonyms in this work according to the context.
Every compound hash function consists of $ k $ atomic hash functions $ h $, each $ h $ being separately initialized with a vector $ \vec{a} = \{a_j\} \in \mathbb{R}^d $ containing independent elements drawn from the normal distribution ($ a_j \sim \mathcal{N}(0,1) $; green boxes in h(x)).

To determine retrieval candidates (\autoref{fig:model}.2), $ h $ calculates the dot product of 
every time step $ \vec{x}_i $ in the time series window with $ \vec{a} $ 
to produce $ X=\{ \vec{x}_i \in \mathbb{R}^d\} $,
thus merging $ d $ tracks in $ X $ to one univariate hash code of length $ t $.
A projection collision for a time step occurs when $ | h(\vec{q}_i) - h(\vec{x}_i) | \le \frac{\omega}{2} $ holds with a given error band $ \omega $. 
Further, a hash collision under an $ h $ happens if the number of projection collisions between the query $ Q $ and $ X $  exceeds a user-defined threshold $ t_s < t $, i.e., for a sub-sequence of the time series.
We can make a stronger similarity claim if the hash collision between $ Q $ and $ X $ happens for all $ k $ atomic hash functions in a compound hash function. 
Finally, $ X $ is considered to be a similarity candidate to $ Q $, when they collide under at least one of the $ l $ compound hash functions (\autoref{fig:model}.2 middle).

Given \ac{lsh}'s probabilistic traits, a non-deterministic number of candidates can end up in the same bucket as the query (arrow \autoref{fig:model}.2 to \autoref{fig:model}.3). 
The mindful reader can experience this aspect through the counter-intuitive results in our performance experiment in \autoref{subsec:quantitative_comparison}.

\toolname extends on Yu et al.'s conventional \ac{mts} modeling scheme\cite{ChenyunYu.2019} by exploiting the univariate hash code of each retrieval candidate. This conceptual extension has two advantages: (1) Rather than calculating the similarity on the high-dimensional dataset, we compute an approximate distance on the low-dimensional hash using any arbitrary distance metric (\autoref{fig:model}.3), resulting in a speedup from linear to constant time with respect to the number of tracks. And (2), we can react to the user wishes since we can adapt the generated hash functions according to the trained user labels, as we will describe in \autoref{subsec:modellearning}.
Although the model building appears to be complex, it can be computed instantaneously and allows for rapid adaption (\R{1}, \R{2}). 



\vspace{-0.3em}

\textbf{Retrieval Invariances:}
Our modeling scheme allows us to account for a range of invariances, which has not only an impact on retrieval performance (\R{3}) but also improves high-level understanding (\R{5}) and similarity perception (\R{2}), as recently demonstrated in\cite{DBLP:journals/tvcg/GogolouTPB19}. 
While horizontal translational invariance is covered by design through the window-based approach (\autoref{fig:model}.1), we are inheriting its weakness of not being able to capture signal length invariance, i.e., longitudinally stretched or compressed patterns. 
Our approach of using a single univariate hash allows us to handle a special type of invariance occurring in \ac{mts} data: track-based invariance. 
To capture distorted/wrapped patterns, i.e., locally morphed shapes, we can even employ an expensive \ac{dtw} calculation for the few candidate windows in the same bucket as the query (upper bound; \autoref{fig:model}.3).
Amplitude scaling invariance, i.e., compressed or enlarged patterns in the y-axis, is implemented by normalizing across all windows in the first modeling step (\autoref{fig:model}.1), as it is typically done in window-based methods.

\vspace{-0.2em}
\subsection{Representative Sampling and Relevance Feedback}
\label{subsec:reprefeedback}

We initiate the model steering in our conceptual model (\autoref{fig:conceptual_model}) through the representative selection and relevance feedback steps.
These steps allow users to incorporate their own domain experiences and subjectivity and give nuanced feedback on how the application should alter its current state. 
This process has two components: (1) it is inherently a visual-interactive (interface) problem (\R{4}, \R{5}) and will therefore be elaborated in \autoref{sec:visual_interface}, and (2) we need to decide which \emph{influential} samples to show to the user. 
In \toolname, the user can give feedback on two different entities: \textit{samples} and \textit{hash tables/classifiers}.

\textbf{Sample Relevance Feedback:} In terms of (un-)classified windows, we can easily expect hundreds or thousands of windows in a database. The user can not process more than maybe a few dozen without being stuck in a tedious labeling process. 
To follow the central idea of (visual) active learning\cite{DBLP:journals/tiis/DudleyK18, DBLP:conf/iui/ArendtSWVD19, dennig2019fdive}, we choose representatives based on two concepts: \textit{exploitation} and \textit{exploration}. Given a query pattern and (initial) feedback from the user, \toolname will exploit this feedback to choose archetypal representatives.
This allows the user to make further distinctions and refinements between the (so far classified) "similar" windows (\R{2}). We, therefore, select the top-k results (user parameter; $1 \leq k \leq 5$) from each hash table.

But, if we only restrict ourselves to showing windows that are deemed similar, we will cause an increasing learning bias, as the user \textit{can not diverge} from the current classifier understanding (\R{6}). To allow for guided exploration, we draw random sample windows from all classifiers, i.e., the average distribution of all individual hash tables.

\textbf{Classifier Relevance Feedback:} As the number of classifiers is limited ($l$ and $k$ are usually lower than 5) showing a overview representation of individual classifiers is possible. The model itself already has a simple representation of the classifier, namely a histogram of similarity buckets. Representing the classifier is thus a projection of that part of the model. 
We have to interpret the potentially occurring histogram patterns in the following manner: A positively skewed shape
$\vcenter{\hbox{\includegraphics[height=1em]{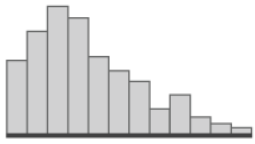}}}$
points to a hash table in which many windows are similar to the query, while a negatively skewed shape 
$\vcenter{\hbox{\includegraphics[height=1em]{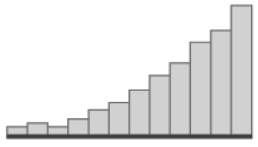}}}$
means the opposite. Unitary 
\raisebox{-.1\height}{\includegraphics[height=1em]{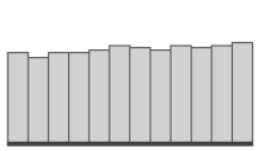}} 
or bi-modal distributions 
\raisebox{-.15\height}{\includegraphics[height=1em]{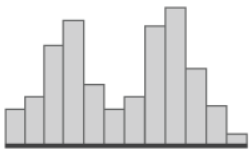}} 
refer to undecided, respectively uncertain, classifiers.

\vspace{-0.5em}
\subsection{Model Learning}
\label{subsec:modellearning}

Conventional \ac{lsh} does not contain trainable parameters. Thus, it cannot be optimized directly by relevance feedback. 
The randomly initialized parameters in the hash functions, however, do not need to be set once and for all.
As described in \autoref{subsec:modelmts}, a hash function $ h $
calculates the dot product $ \vec{a} \cdot \vec{x}_i $ with its hashing vector $ \vec{a} $ and i-th time step in window $ X=\{\vec{x}_i \in \mathbb{R}^d\} $.
This operation can be interpreted as a weighted merge of $ d $ tracks in $ X $ with the track weights in $ \vec{a} $.

To capture the relevance feedback (\R{2}), we propose to modify $ \vec{a} = \{a_j\} $ with a learned weighting vector $ \vec{w} = \{w_j\} $ and use this preference-adapted parameter vector $ \vec{b} = \vec{w} * \vec{a}  $
for the similarity calculation.
To avoid vanishing and exploding parameters, we want to retain the expectation of the magnitude of $ \vec{a} $, namely $ E(||\vec{b}||) = E(||\vec{a}||) $. 
This is achieved by normalizing the magnitude of $ \vec{w} $ to $ \sqrt{d} $. 
We keep detailed mathematical proof in the Appendix. 

We allow two types of relevance feedback in \toolname (\R{4}), namely feedback on the classifiers (hash tables) and on the result samples. Consequently, we maintain two weight vectors $ \vec{w}_c $ and $ \vec{w}_s $ for classifiers, respectively samples, to track their importance adoption.

\textbf{Classifier Relevance Adaption:} The feedback on classifiers can be implemented in a straightforward manner.
Let $ A = \{\vec{a}_{pos}\} $ be the parameter vectors of all hash functions labelled as positive by the user.
Then, we can define $ \vec{a}_c = \{a_{c,j}\} = \{\sum_{\vec{a}_{pos} \in A} a_{pos,j}^2\} $.
After normalization to magnitude $ \sqrt{d} $ we get 
$ \vec{w}_c = \vec{a}_c \frac{\sqrt{d}}{||\vec{a}_c||} $

\textbf{Sample Relevance Adaption:} The feedback on the positively labeled samples, 
$ \mathbb{C}^{+} = \{C_{pos}\} $,
is transformed to a per-track importance. For this, we make use of the \ac{dtw} distance between the tracks of the positively labeled samples and the query $ Q $.
Let 
$ \vec{q}_j $ be the j-th track of $ Q $, $ \vec{c}_j $ the j-th track of 
$ C_{pos} $. 
We can define $ \vec{z}_j $ as the aggregate distance between $ \vec{c}_j $ and $ \vec{q}_j $ with
$ \vec{z}_j = (\sum_{C_{pos} \in \mathbb{C}^{+}} DTW(\vec{q}_j, \vec{c}_j))^2 $. 
Next, we normalize the entries in $ \vec{z}_j$ between $ [0,1] $ with $ \vec{z}^* = \{z_j^*\} = \{z_j / \sum_{j=1}^{d}z_j\} $.
Converting distance to its negatively correlated weight vector
yields $ \vec{w}_s^* = \{1-z_j^*\} $,
which is subsequently normalized to 
$ \vec{w}_s = \vec{w}_s^*\frac{\sqrt{d}}{||\vec{w}_s^*||} $

We are unifying both vectors $ \vec{w}_c $ and $ \vec{w}_s $ into $ \vec{w}_r $ through a linear combination with the previous feedback round values $ \vec{w}_{r-1} $, $ \vec{w}_{c,r} $ and $ \vec{w}_{s,r} $ followed by a normalization (\autoref{eq:weight}) with a learning rate $ \alpha $, which can be gradually modified to enforce exploration stability\cite{M.Behrisch.2014}. 

\begin{equation} 
\begin{aligned}
\vec{w}_r^\ast & = (1-\alpha)\vec{w}_{r-1} + \frac{\alpha}{2}(\vec{w}_{c,r} + \vec{w}_{s,r}) \\
\vec{w}_r & = \vec{w}_r^\ast \frac{\sqrt{d}}{||\vec{w}_i^\ast||}
\end{aligned}
\label{eq:weight}
\end{equation}

\noindent\textbf{Query Specificity Adaption:} As shown in the second arrow in \autoref{fig:pipeline}, we not only redefine the \ac{lsh} hash functions, but also the search pattern. 
In other words, we actively update the query for specificity based on the received user feedback. 
In each training iteration, we integrate $ Q $ (search query) and $ \mathbb{C}^{+} $ (set of positively labeled samples). 
This operation is known to be non-trivial\cite{F.Petitjean.2014} and has an impact on future exploration direction.
After the trial with the na\"ive elementwise average yields unsatisfactory results, because the average often resembles none of the original windows,
we decided for \ac{dba}\cite{FrancoisPetitjean.2011}, as it takes distortion and time shift into consideration\cite{FrancoisPetitjean.2011,F.Petitjean.2014,G.Forestier.2017}

\section{Interactive and transparent visual interface for \acl{mts} retrieval}
\label{sec:visual_interface}

\begin{figure*}[ht]
    \centering
    \includegraphics[width=\linewidth]{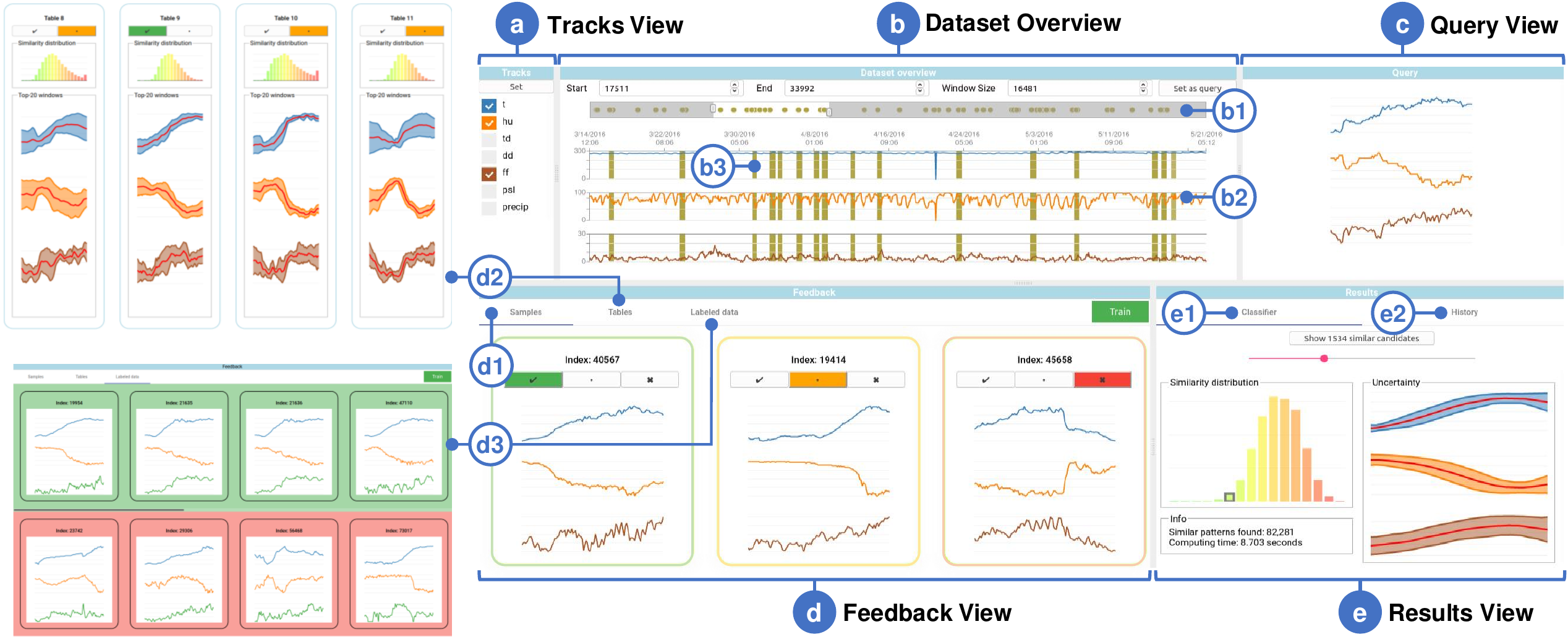}
    \caption{
        \textbf{Interface Overview:} We plot the selected tracks from the \textit{Channels View} (a) in the \textit{Dataset Overview} (b) along with the user-defined query in (c).
        The \textit{Feedback View} (c) depicts classification results samples and information about the classifiers (hash tables), and is used to receive relevance feedback to update the model.
        The \textit{Results View} (e) displays the result distribution and provides exploration state management. 
}
    \label{fig:interface_overall}
\end{figure*}

Since capturing and processing human subjectivity is one of the central goals of this project (\R{2}), and to test the \toolname's functionality in a realistic (\R{6}) and (also for us) understandable environment (\R{5}), we designed the visual interface depicted in \autoref{fig:interface_overall}.
It consists of five interlinked views facilitating querying, relevance feedback, exploration and transparency, and result overview.
The web-based user interface is connected to the \toolname backend algorithm through a REST API.
Both backend (primarily Python) and frontend code (mostly Angular) are detailed and freely available under \codeurl{}.

\subsection{Dataset Overview, Channels View, and Query View.} 
\label{subsec:views_above}

The \textit{Dataset Overview}, \textit{Channels View}, and \textit{Query View} collaborate closely.
The \textit{Dataset Overview} (\autoref{fig:interface_overall}.b) plots tracks selected in the \textit{Channels View} (\autoref{fig:interface_overall}.a) together with the window labels (\autoref{fig:interface_overall}.b3). 
Besides the direct panning and zooming in the plot, a range slider (\autoref{fig:interface_overall}.b1) above the track curves (\autoref{fig:interface_overall}.b2) serves as a mini-map for navigation and overview of labeled windows. 
Real-life datasets easily reside over millions of data points, posing a challenge for a fluent interaction and clean visualization design on a screen with limited resolution.
While algorithmic, e.g.,\cite{douglas1973algorithms}, or visualization-focused, e.g.,\cite{peng2008method}, solutions exist that compress data with information loss, we dynamically downsample each track to hundreds of data points during the zoom and increasingly add information to the coarse overview as the user zooms in.

The \textit{Query View} (\autoref{fig:interface_overall}.c) shows the user-defined multi-track query.
Our query is defined in a query-by-example manner by selecting a region in the \textit{Dataset Overview} (\R{4}, \R{6}). 
The user is allowed to change the query on-the-fly at any time, but we have to repeat the hashing process whenever the \emph{query size} changes, in other words, if the window width differs. Typically a user has to wait around 8-10 seconds for the model rebuilding on a 100K 100-track, 100 timestep window dataset. While not being in the focus of this project, we found that our query definition interface lacks the option to define pattern sequences across tracks. We plan to investigate better query interfaces for \ac{mts} data following recent examples, such as\cite{DBLP:conf/vizsec/CappersMEW18}.

\subsection{Feedback View}
\label{subsec:feedback_view}
The \textit{Feedback View} (\autoref{fig:interface_overall}.c) shows representatives of the classifier result, visualizes hash tables, and keeps track of labeled data (\R{5}). We differentiate three respective tabs for different purposes:

The \textit{Samples Tab} (\autoref{fig:interface_overall}.d1) lists representatives or samples of the classified windows.
The windows are surrounded by frames color encoded from green to red, indicating declining similarity predicted by the classifier.
Right above the curves, the user is invited to label the window by clicking
$\vcenter{\hbox{\includegraphics{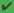}}}$
for similar, 
$\vcenter{\hbox{\includegraphics{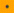}}}$
for indecisive and 
$\vcenter{\hbox{\includegraphics{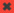}}}$
for dissimilar, in order to modify the sample relevance, as described also in \autoref{subsec:modellearning} (Sample Relevance Adaption). 

\toolname visualizes hash tables in the \textit{Tables Tab} (\autoref{fig:interface_overall}.d2).
The hash tables are depicted as histograms showing the similarity distribution of one hash function. Each bin corresponds to a bucket in the corresponding hash table. 
The bars in the histograms are color encoded from green to red, indicating the decreasing similarity to the current query.
To better understand how well our hash functions work, we plot the mean, minimum, and maximum values of each time step among the top-20 similar windows for each hash table. 
The mean values portray the pattern shape as perceived similar by the hash function,
whereas the minimum and maximum values form a lower and upper bound approximation of the pattern variance.
The variance band's tightness implies the (un-)certainty or importance of the track during classification.
Based on this visual encoding, the user can modify the hash tables' importance by clicking
$\vcenter{\hbox{\includegraphics{figures/sec_4_visualinterface/interface_labels/feedback_positive.png}}}$
for important and 
$\vcenter{\hbox{\includegraphics{figures/sec_4_visualinterface/interface_labels/feedback_neutral.png}}}$
for indecisive, as described also in \autoref{subsec:modellearning} (Classifier Relevance Adaption).

The labeled windows for the current round are tracked in the \textit{Labeled Data Tab} (\autoref{fig:interface_overall}.d3). The user can revise and modify the decisions before clicking the 
\fboxrule-2pt\fboxsep2pt
\colorbox{traingreen}{\begin{minipage}{2.8em}
\centering
  \texttt{Train}
\end{minipage}} button (\R{5}). 
The hash table and label user feedback will be considered during the next training round. 

\subsection{Results View}
\label{subsec:result_view}
The \textit{Results View} (\autoref{fig:interface_overall}.e) shows the classification outcome for all windows and provides a visual undo/redo panel as a tree visualization to revert entire model learning decisions (\R{5}); Figure omitted. 

In the \textit{Classifier Tab} (\autoref{fig:interface_overall}.e1), we show a result histogram to depict the similarity distribution of all windows to the query. 
Clicking a bin in the histogram results in the reconstruction of the visual pattern prototype analogous to the curves in the \textit{Feedback View}.
Rather than using the top-20, this pattern result view summarizes all windows in the chosen bin.
The mean curves also show the average form within this bin, 
bounded by each time step's minimum and maximum values to illustrate the variance within tracks.
The histogram and the reconstructed shape help the user better understand the classification result and guide the required strictness during the labeling process (\R{3-5}). 
One central interaction in this panel is that the user can set the number of top candidates and display them in the data overview. 

\section{Evaluation \toolname on Performance and Usability}
\label{sec:evaluation}

\begin{figure*}[ht]
    \centering
    \begin{subfigure}[b]{0.30\linewidth}
        \includegraphics[width=\linewidth,trim=15 15 10 5,clip]{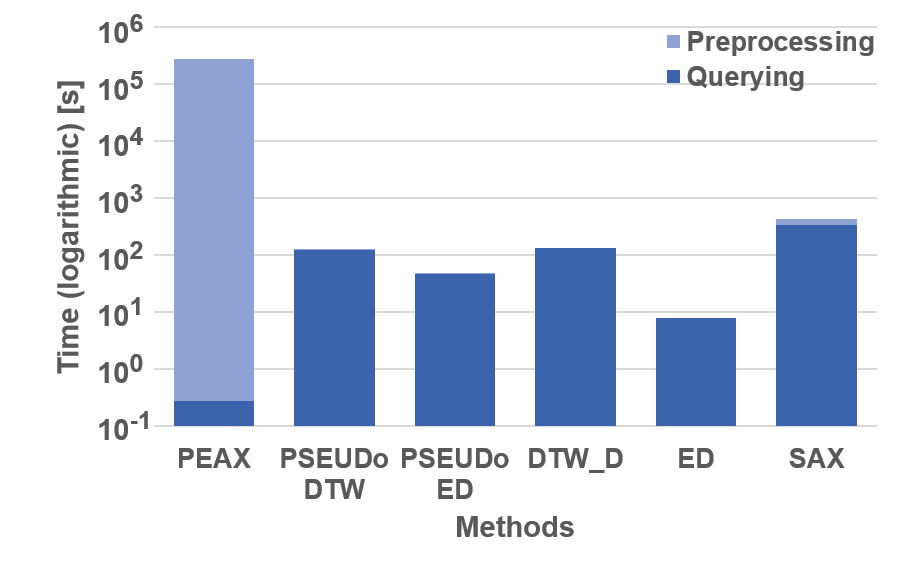}
        \caption{Overall processing time comparison.}
        \label{fig:peaxcomparison}
    \end{subfigure}
    \hfill
    \begin{subfigure}[b]{0.31\linewidth}
        \includegraphics[width=\linewidth,trim=15 15 10 5,clip]{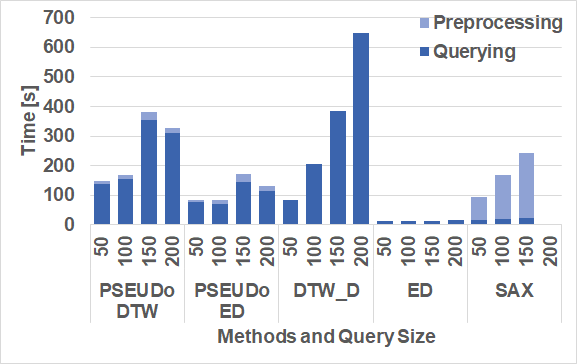}
        \caption{Processing time vs. query size for different methods.}
        \label{fig:speed_query_size}
    \end{subfigure}
    \hfill
    \begin{subfigure}[b]{0.34\linewidth}
        \includegraphics[width=\linewidth,trim=15 15 10 20,clip]{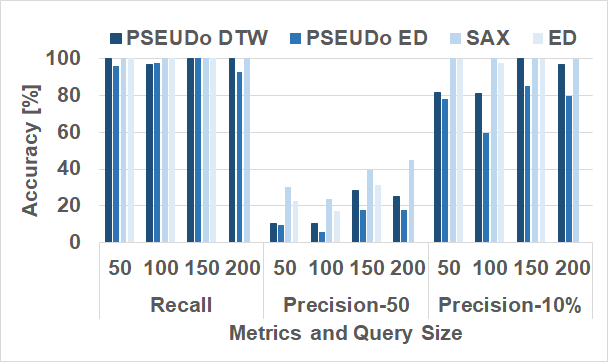}
        \caption{Accuracy vs. query size across different techniques.}
        \label{fig:accuracy_query_size}
    \end{subfigure}
    \caption{\textbf{Performance charts} for a selection of three out of eight experiments that we conducted to validate \toolname's performance to the state-of-the-art TS retrieval algorithms \acfi{dtwd}, \acfi{ed}, \acfi{sax}, as well as PEAX\cite{Lekschas.2020}. \toolname outperforms autoencoder-based systems, like PEAX, by a factor of 5550x in its fastest variant \toolname-ED.}
    \label{fig:test}
\end{figure*}

We evaluated \toolname against the requirements described in \autoref{subsec:requirements_for_multi_track_time_series_exploration} through three distinct evaluation threads. 
First, a quantitative/technical evaluation showing \toolname's advancements in terms of \textbf{R1} (scalability), \textbf{R3} (accuracy), and \textbf{R4} (steerability) in comparison to state-of-the-art pattern retrieval techniques.
Second, a visual quality inspection that deliberately takes the user's viewpoint and visually demonstrates the pattern coherence without relying on potentially insufficient (visual) quality metrics\cite{DBLP:journals/cgf/BehrischBKSEFSD18}
And third, a case study that outlines usability aspects and how a target user would apply \toolname's visual interface.

We used five \ac{mts} datasets of varying complexity, namely 
a gas sensor dataset (1 TS/ 4.2 million time steps/ 16 tracks), an EEG dataset (1/0.9M/70), a synthetic dataset (10K/120/3), a meteorological dataset (1/21.9M/5) and an epigenomic dataset (1/120K/1).
The gas sensor dataset was also used by Yu et al. to showcase their \ac{lsh}-based algorithm\cite{ChenyunYu.2019}. 
We chose the EEG dataset since it contains a large number of tracks.
Finally, we added PEAX to the comparison with its original single-track epigenomic dataset\cite{Lekschas.2020}. 
All technical experiments were conducted on a standard laptop running on 64-bit Ubuntu 20.04 with Intel i5-8350U CPU, 16GB RAM, and 512GB SSD.
More about these datasets and our experiments can be found in the Appendix
and on our website.

\subsection{Quantitative Comparison}
\label{subsec:quantitative_comparison}

We evaluated \toolname by comparing it to the state-of-the-art techniques \ac{ed}, \ac{dtwd}, and \ac{sax}
on the gas sensor and EEG dataset.

The scalability characteristics of \toolname are demonstrated by measuring the computation time on increasing data volume. 
We distinguished between \emph{preprocessing} and \emph{querying time}. 
For our accuracy investigation, we rely on precision and recall. We compute the top-50 similar windows as suggested by \ac{dtwd}\cite{MohammadShokoohiYekta.} (referred to as $ W $) and treat them as ground truth.
We chose this approach because \acs{dtw}-based techniques, although being highly inefficient, are currently considered the best in terms of accuracy for time series classification\cite{Bagnall.2017}.
Aware of the risk that \ac{dtwd} might perform poorly, we furthered the accuracy evaluation with a visual quality inspection in \autoref{subsec:visual_quality_inspection}. 
Recall is defined as the portion of $ W $ also classified as similar by the examined technique.
For precision, we defined a strict metric called precision-50 as the portion of the top-50 windows classified by the examined technique that can also be found in $ W $. 
At the same time, we define a relatively looser metric, called precision-10\%, as the portion of the top 10\% similar windows of the examined technique that are also in $ W $.

\textbf{Experiment Overall Speed:}
The primary claim of this paper is that we can beat the model building and inference time of autoencoder-based methods, such as PEAX\cite{Lekschas.2020}, while still serving similar objective and subjective retrieval performance (\R{1}, \R{3}). 


As \autoref{fig:peaxcomparison} depicts, \toolname is 2167x faster with its slowest but most precise instantiation \toolname-DTW and 5557x faster with \toolname-ED compared to PEAX (277,200s preprocessing/0.279s querying) on the same 120kb DNase-seq dataset that was used in the PEAX evaluation\cite{Lekschas.2020}. 
\toolname (3s/124s) is only seemingly on par with \ac{dtw} (0s/135s) as its efficiency boost mainly surfaces for many tracks.

\textbf{Experiment Query-Size vs. Speed/Accuracy:}
This experiment evaluated the query size's influence on scalability and accuracy (\R{1}, \R{3}). We randomly picked ten queries from the gas sensor dataset and measured accuracy and elapsed time with varying query sizes (50/100/150/200 time steps per window). We kept \toolname's relevance feedback switched off in favor of \ac{lsh}'s objective performance.
We illustrate our results in \autoref{fig:speed_query_size} for speed and \autoref{fig:accuracy_query_size} for accuracy. 


\autoref{fig:speed_query_size} reveals the good scalability characteristics of \toolname-DTW for large query sizes (150/170/380/320 seconds), especially compared to \ac{dtwd}.
\toolname-\ac{ed} has roughly half the processing time as compared to \toolname with \ac{dtw} without considerable accuracy loss. 
Although \ac{sax} features low processing time (95/170/240/-s), its large memory consumption impedes its scalability.
Surprisingly, \ac{ed} outperforms all competitors in computing time and accuracy on the gas sensor dataset used in this experiment.
On a side note, \toolname needs less time for query size 200 than for 150 because the number of candidates that end up in the same bucket as the query is non-deterministic, as explained in \autoref{subsec:modelmts}


\autoref{fig:accuracy_query_size} shows a generally better recall than precision, especially for \toolname. Although showing many false positives due to its stochastic nature, \toolname indicates a good similar pattern retrieval performance, particularly here without relevance feedback. 
The result seems to reach its optimal around the query size of 150, presumably because this pattern length is characteristic in the EEG dataset. 
\toolname-\ac{dtw} slightly outperforms \toolname-\ac{ed}, most likely because the ground truth is generated with \ac{dtwd}.
Overall, this experiment does not allow us to judge which technique is better in terms of accuracy but instead serves the purpose of confirming a comparable accuracy of these methods. 

\textbf{Experiments Dataset-Size vs. Speed/Accuracy:}
The second experiment varied dataset size (25K/50K/75K/100K) instead of query size on the EEG dataset. 
For space reasons, its detailed result charts can be found in the Appendix.
Generally, we see a relatively stable Precision-10\% accuracy for \toolname-DTW (89.2/93.5/96.6/97.1\%), \toolname-ED (84.3/95.3/97.5/96\%), \ac{sax} (100/100/100/-\%), \ac{ed} (100/100/100/100\%). 
Please note that \ac{dtwd} was used to generate the baseline. 
On the other hand, our processing time charts show a roughly linear time complexity increase across the dataset sizes for all techniques; \toolname-DTW (24/44/71/91s), \toolname-ED (17/30/44/59s), \ac{dtwd} (45/88/133/176s), \ac{sax} (30/58/92/-s), \ac{ed} (2/4/6/9s).
\toolname outperforms \ac{dtwd} thanks to sublinear complexity for preprocessing and lower constant factor for querying.
Due to a memory problem, \ac{sax} is missing for the 100K experiment.
Compared with \ac{dtwd}, \toolname-\ac{ed} again accelerates querying by a factor of two with comparable accuracy.

\textbf{Experiments No. Tracks vs. Speed/Accuracy:}
This experiment examines the impact of the number of tracks on performance. 
We used the EEG dataset in this experiment and varied the number of tracks by 20/40/60. 
Our results for this experiment are conclusive with the previous experiment and depicted in the Appendix, as well.
We see that \toolname's sublinear complexity is especially beneficial as the number of tracks increases. \toolname-DTW (65/69/75s) and \toolname-ED (35/40/57s) show only minimal processing time increase across the different number of tracks as opposed to \ac{dtwd} (258/318/400s). \ac{ed} (5/6/8s) is again the positive outlier but will most likely not meet the user's accuracy standards in real-world scenarios. \ac{sax} is abscent in the plot due to memory issues. In terms of accuracy, we can see also comparable results to Experiment 2 in terms of Precision-10\%: \toolname-DTW 96.8/90.7/83.8\%), \toolname-ED (73.2/86.7/81\%), \ac{ed} (100/100/100\%).

\textbf{Experiment Steerability:}
We evaluated
\toolname's \textbf{steerability} by demonstrating the retrieval convergence of our algorithm. 
By manually assigning a fixed weight vector $ \vec{w} $ in each feedback round, 
we should notice that the strongly 
weighted tracks become more similar to the query, and their variance bands tighten.
We experimented with five random queries and put increasingly more weight on \textit{Track 1} while doing the opposite for \textit{Track 3}.
We plot the mean, minimum, and maximum curves of the tracks in the top-50 windows found by \toolname.
Finally, as a subjective measure, we report the \acs{dtw}-distance between the corresponding tracks in the query and the times series windows as a progression over four feedback rounds.

\begin{figure}[h]
    \centering
    \begin{subfigure}[b]{0.4\linewidth}
        \includegraphics[width=\linewidth,trim=3 3 0 0,clip]{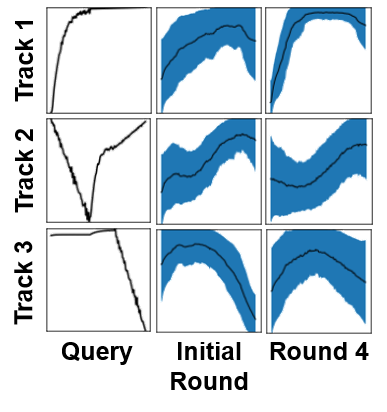}
        \caption{Prototypes and variance bands.}
        \label{fig:steerability_prototype}
    \end{subfigure}
    \begin{subfigure}[b]{0.54\linewidth}
        \includegraphics[width=\linewidth,trim=3 3 3 5,clip]{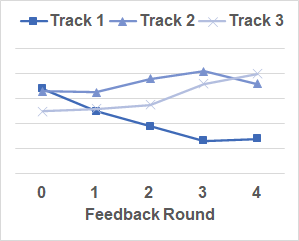}
        \caption{Impact of feedback: distance to query}
        \label{fig:steerability_line_chart}
    \end{subfigure}
    \caption{\textbf{Steerability evaluation.} (a) Avg. shapes and variance bands for 3 tracks before/after training (prototypes). Track 1 received positive feedback and became increasingly similar to the query; its variance band narrowed. Track 3 received negative feedback and \toolname reacted conversely. (b) The track-wise \ac{dtw} confirms \toolname's steerability.
    }
    \label{fig:steerability}
\end{figure}

\begin{figure*}[ht]
    \centering
    \includegraphics[width=\linewidth]{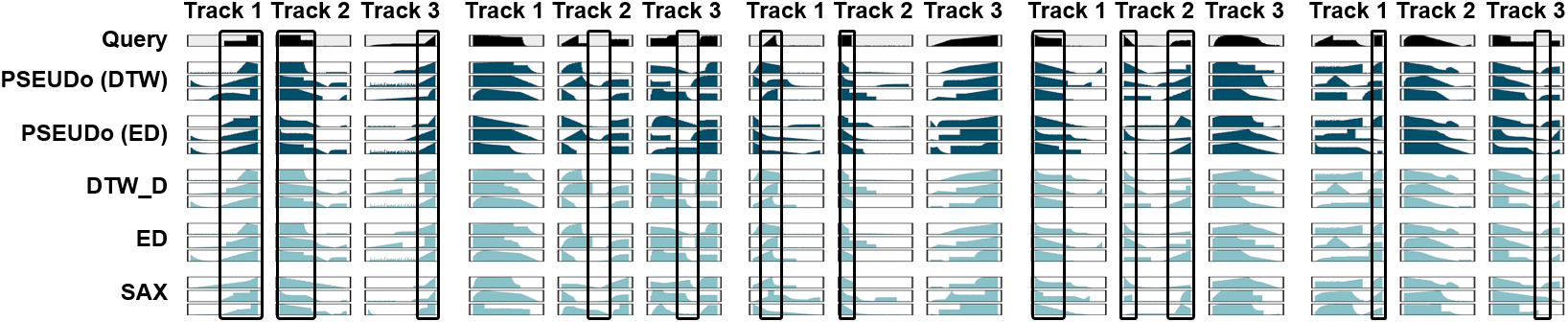}
    \caption{Visual comparison of 3-nearest neighbours for five 3-track queries and five retrieval techniques. Black boxes mark important features.}
    \label{fig:visual_quality_inspection}
\end{figure*}

As indicated in \autoref{fig:steerability_prototype}, 
the strongly weighted \textit{Track 1} 
exhibited less variance and became increasingly similar to \textit{Track 1} in the query compared to the shape before training. 
The opposite happened for \textit{Track 3}.
This observation is consistent with the decreasing \acs{dtw}-distance between the \textit{Track 1} windows and the query, respectively, the increasing \acs{dtw}-distance for \textit{Track 3} (\autoref{fig:steerability_line_chart}).
For the sake of generalizability, we show repetitions of these experiments with other queries in the Appendix.
These experiments show the same behavior.
Consequently, we can conclude that the result classified by \toolname can be steered by means of a weight vector.

\subsection{Visual Quality Inspection}
\label{subsec:visual_quality_inspection}

Since the quantitative comparison used the result of \ac{dtwd} as ground truth and thus implied absolute correctness, we felt compelled to follow the argument that no single similarity measure accounts for human judgement of time series similarity\cite{M.Correll.2016b} and substantiate our evaluation with a visual quality inspection.
We chose five interesting patterns and visually inspected the five top hits with \ac{dtwd}, \ac{ed}, and \ac{sax}, while keeping \toolname's relevance feedback mechanism switched off.
The comparison to PEAX had to be omitted because it does not support multiple tracks readily.
This time, we used the synthetic dataset due to its recurrent and easily recognizable patterns and show the results in \autoref{fig:visual_quality_inspection}

We can draw several conclusions: 
(1) \toolname delivers comparable visual quality to state-of-the-art techniques;
(2) No method significantly outperforms the others, and the results are visually similar.
Considering time and space consumption, \toolname-\ac{ed} scores the runner-up position after \ac{ed}. 
However, the vanilla version of \ac{ed} does not contain trainable parameters and thus cannot be updated with relevance feedback, though enhanced version exists in\cite{Keogh.1999}.

\subsection{Case Study}
\label{subsec:case_study}

We demonstrate \toolname's usability, especially in light of the requirements (\R{4-6}) from \autoref{subsec:requirements_for_multi_track_time_series_exploration}. We conduct our case study on the MeteoNet meteorological dataset (1 TS / 88,197 time steps/ 7 tracks)\cite{METEOFRANCEdataset}. 
and show the interactive flow and responsiveness of our visual interface 
in the accompanying video in the Appendix.

\begin{figure}[ht]
    \centering
    \begin{subfigure}[b]{0.18\linewidth}
        \includegraphics[width=\linewidth,height=2.5cm]{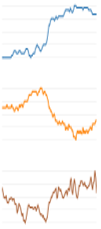}
        \caption{Query}
        \label{fig:initial_query}
    \end{subfigure}
    \begin{subfigure}[b]{0.78\linewidth}
        \includegraphics[width=\linewidth,height=2.5cm]{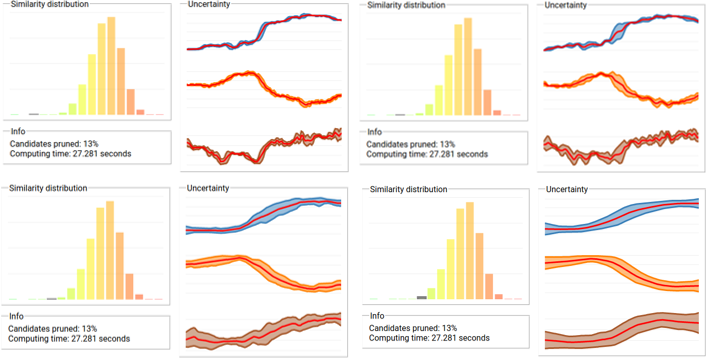}
        \caption{Found patterns in first four bins}
        \label{fig:initial_result}
    \end{subfigure}
    \caption{Initial query and patterns found after the first feedback round.}
    \label{fig:inital_query_result}
\end{figure}

Imagine Tom, a meteorologist who investigates the precursors of tornado formation in rural areas. This is certainly a rare and highly time-dependent event in which several weather phenomena have to come together.
Tom loads the weather station data in \toolname and checks boxes for \textit{temperature}, \textit{humidity}, and \textit{wind speed} as shown in \autoref{fig:interface_overall}.a, because he believes that the interesting periods correlate with an increase in temperature, falling humidity values, and some unknown behavior of the wind speed in the early stages of the formation process.
He browses the \textit{Data Overview} (\autoref{fig:interface_overall}.b) and finds a pattern shown in \autoref{fig:initial_query} that seemed to fit his early hypothesis that the humidity drastically drops before the wind speed increases.
He pins this pattern as a query and starts the initial search. 
After approximately nine seconds, i.e., the model building time of \toolname, he notices that the \textit{feedback view} and the \textit{result view} are packed with information.
He eagerly checks the result view in \autoref{fig:inital_query_result} and notices that the result has not captured his desire to a satisfactory extent. 
Because (1) there are not many windows in the first bins of the result histogram and (2) the windows in the first bins are largely overfitting the query too much, a general problem in most query-by-example retrieval systems. 
This is indicated by the tight variance bands around the prototypes, as well as the quick rise and fall of the curves in the prototypes that overly resembled the query. 

\begin{figure}[ht]
    \centering
    \begin{subfigure}[b]{0.18\linewidth}
        \includegraphics[width=\linewidth,height=2.5cm]{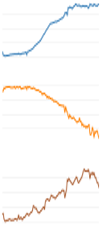}
        \caption{Query}
        \label{fig:final_query}
    \end{subfigure}
    \begin{subfigure}[b]{0.78\linewidth}
        \includegraphics[width=\linewidth,height=2.5cm]{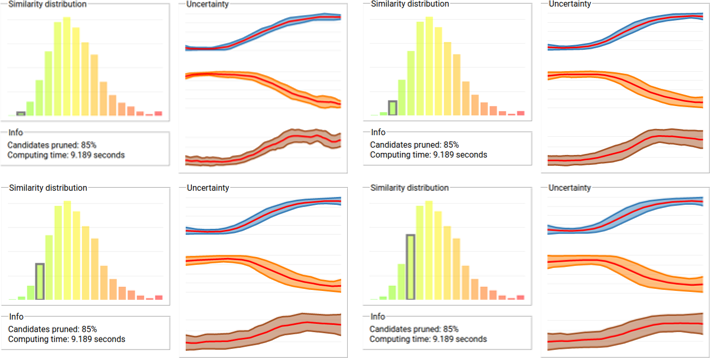}
        \caption{Found patterns in first four bins}
        \label{fig:final_result}
    \end{subfigure}
    \caption{Query and found patterns after three feedback rounds.}
    \label{fig:final_query_result}
\end{figure}

Next, Tom expresses his subjective similarity perception, i.e., his accepted range of fuzziness for his mental model of tornado formation, along with a more precise definition of the query, i.e., his gained knowledge from the first labeling iteration. 
He first marks sampled windows under the \textit{samples tab} in the feedback view, tagging a smoother temperature curve, respectively humidity decrease, as shown in \autoref{fig:interface_overall}.d1. Tom labels crudely hash functions that match his expectations (shown in \autoref{fig:interface_overall}.d2) in the \textit{tables tab}. 
Then, after a quick validation of the labeled windows in the \textit{labels tab}, he clicks the "Train" button to adapt the model. 
After again around eight seconds, the result succeeding the first training iteration can be explored. 
He notices a visual improvement and conducts one more feedback round. 
The final result is shown in \autoref{fig:final_query_result}. 

Not only do the prototypes better match his expectation about the early stages of tornado formation, but also the query specification improved each time through our \ac{dba} approach described in \autoref{subsec:modellearning}.
He is content with the search result at this point and focuses now only on his result presentation (\autoref{fig:interface_overall}.e). 
By interactively checking the result histogram, he determines a cut-off point between similar and dissimilar windows, i.e., supporting vs. contradicting events in the dataset. 
The similar windows was refreshed automatically in the data overview giving even more insight into (until now) neglected weather features. 
Satisfied with the result, he continues to check the found interesting instances against his available ground truth from, e.g., news articles.

\section{Related Work}
\label{sec:relatedwork}


For didactic purposes, we discussed the gap in the research literature throughout the paper, e.g., in \autoref{sec:background} or \autoref{sec:classifierlearning}. This Section should give the reader a broader context and focus on advancements in algorithmic and interactive \ac{uts} and \ac{mts} data analysis.


\textbf{Automated Analysis:} In the past, automated TS analysis has been tackled mostly with feature engineering-based approaches or through direct distance metrics-based techniques, such as \acf{ed}, \acf{dtw}\cite{ding2008querying}, or \acf{dtwd}\cite{MohammadShokoohiYekta.}.
For example, Single Value Decomposition (SVD) \cite{faloutsos1994fast, weng2008classification}, Discrete Fourier Transformation (DFT)\cite{faloutsos1994fast}, Discrete Wavelet Transformation (DWT)\cite{chan1999efficient} and \ac{paa}\cite{keogh2001dimensionality} all present varyingly complex and efficient problem space transformations that divide TS data into segments and define a real-valued approximation for this segment. But, representing the data using real-values has been shown to lead to algorithmic limitations\cite{lin2012pattern}. \textit{Symbolic} representations, on the other hand, convert the TS into a fixed vocabulary of categorical symbols. 
The most popular symbolic reduction technique is \acf{sax}\cite{Lin.2007}, which uses \ac{paa} and a distribution-based lookup-table to discretize a time series into a series of symbols of a fixed-size alphabet.
Its data-size independent dimensionality reduction and simplicity are the reasons for its popularity and many extensions\cite{lkhagva2006extended, shieh2009isax, senin2013sax}. 
Most recently, Loeschcke et al. presented a progressive parameter exploration to find task-adapted parameter settings for \ac{paa} and SAX\cite{DBLP:conf/eurova-ws/LoeschckeHS20}. 
Although originally created for \ac{uts}, this technique can be extended to \ac{mts}\cite{park2020sax}. 
The interested reader will find more details in the survey papers of Ding et al.\cite{ding2008querying},  focusing on the comparison of problem space transformation approaches and Xing et al.\cite{DBLP:journals/sigkdd/XingPK10} focusing on similarity measures for pattern search in sequential data.

\smallskip

\textbf{Semi-automated Analysis:} Pattern search in \ac{mts} and \ac{uts} analysis is also approached from a human-in-the-loop viewpoint. These works primarily emphasize the effective communication between the domain-knowledge savvy analysts and the algorithmic analysis system. As in \toolname, in almost all research on pattern search in (multivariate) time series data, the query is itself a (multivariate) time series fragment. Usually, these queries are either represented by sketch\cite{sketch1, sketch2, buono2008interactive} or by example\cite{buono2005interactive, Lekschas.2020}. Both have been shown to provide efficient query specifications in mixed-initiative systems\cite{M.Correll.2016b}. In the case of query-by-example, the user chooses a segment of the data as the query, and the retrieval system compares the query to a database in a feature- or data-space\cite{cheng2009detection, li2018anomaly}. 
Hao et al. \cite{motif} use a similar approach for finding frequently occurring patterns in MTS data, the difference being that they put an emphasis on pattern visualization and distinctly visualize all frequently occurring patterns.

\smallskip
\textbf{Visualization Approaches:} While the works above reduce the data overload by only focusing on sections and important subregions of the data, this domain has also been addressed from a visualization-focused angle. For example, Buono et al. emphasise the importance of visualizing the \ac{mts} pattern search process in TimeSearcher 2\cite{buono2005interactive}. 
In this system, the user can select up to eight channels for inspection and zoom in and out to generate a simple overview of the dataset. 
An easy-to-use and effective visualization is demonstrated by Peng\cite{peng2008method}, where each channel of an MTS is discretized into a series of categorical colors and stacked to show the inter-channel correlation. 
Buchmüller et al.'s MotionRugs\cite{DBLP:journals/tvcg/BuchmullerJCBK19} adopt a similar idea to incorporate additional spatial information. 
Bach et al.\cite{DBLP:journals/tvcg/BachSHMGD16} and Bernard et al.\cite{bernard2012timeseriespaths} use a more abstract representation as a path in a 2D space. 
Even more abstract, e.g., glyph-based overview visualizations, are presented in\cite{kaleidomap, pham2019mtsad}. 
Although many of these visualization approaches provide a good overview, they are often highly task-oriented (overview vs. detail) and accept to hide a lot of nuanced information from the user.

\section{Discussion and Future Work}

Compared to the related work on interactive \ac{mts} analysis, our approach goes beyond the state-of-the-art by incorporating three main aspects that make \ac{mts} data exploration more tractable for real-world applications. 
Firstly, our system offers an adaptive, as opposed to static one-shot classification, making it a user-centric Visual Analytics approach. Secondly, it relies on one of the most scalable data processing techniques ever invented: hashing-based algorithms. Thirdly, the notion of ``buckets'' can be easily understood, allowing \toolname to be applied in less ML-savvy application environments.

We also came across implementation/design decisions and challenges during the research project that we would like to discuss here. 
On the conceptual side, we found that a thorough task taxonomy for \ac{mts} data is missing. 
While we can still map out \toolname's high-level tasks into Brehmer and Munzner's typology\cite{DBLP:journals/tvcg/BrehmerM13}, e.g., we serve \texttt{browse}, \texttt{explore}, \texttt{locate}, and \texttt{lookup} (all search-related) tasks, the specific \ac{mts} tasks like finding
patterns with significant cross-track time shifts
or tasks that assess the invariance properties of specific patterns are apparently on a different abstraction level. 

We decided early on in the project to focus our attention on the backend aspects rather than the frontend. One can see this aspect distinctively in the rather simplistic use of standard visualizations and the implemented query-by-example system, which work flawlessly and effortlessly for the intended tasks. 
In the future, however, we plan to extend in two directions: First, we will tackle challenging questions, like \textit{How can we let a user specify a) multi-track queries or b) queries with a temporal relationship between them?} with new query definition panels and plan to apply interactive augmentation, like in Shadow Draw\cite{DBLP:conf/vissym/ShaoBSLSBK14}, to help with this process. Second, we will expand our interface with one additional view that should work in unison with the resulting histogram: A SOM-based pattern space overview, such as presented in\cite{DBLP:journals/tvcg/SachaKBBSAK18}. 
This view can be easily understood by users and proves to display (automatically) visual patternclasses into distinct visual space regions. 

Currently, our approach is following the thought that \ac{mts} is interpreted as multi-dimensional \textit{numeric} vectors in sequential order. We have to state that we are independent of the so-called resolution of the time series. However, there are other application areas where such a simplistic \ac{mts} interpretation does not hold, and the entries in distinct dimensions can be categorical, ordinal, or even a complex data type. For example, in crime analysis, one dimension could be a surveillance webcam frame, while the other ones are numeric features. \toolname's design is flexible enough to learn distinct (time-synchronized) models in parallel to tackle these challenges. 

One challenge for which we do not have a good solution yet is \textit{biased search}. Right now, we include negative and indecisive labels to promote the target class separation, adding to the confirmation bias in every iteration. One approach implemented in \toolname is to suggest enormously different samples to give the user an option to ``broaden'' the search. 
Nonetheless, we can regard every new positive label distinct enough from the current set of already labeled items as a new exploration thread/fork. These complex analytic provenance concepts and potentially task-change metrics still need to find their way into future VA systems.

On the technical side, we plan to experiment with a multi-scale LSH variant\cite{DBLP:conf/mir/WengJSSCA15} to address the problem of introducing a user context switch by having to retrain our models when the query size changes. We plan to enhance \toolname's feedback-driven mechanism with better active learning suggestions and extend \toolname to other time series analysis problems, like anomaly detection and motif discovery.

\section{Conclusion}
\label{sec:conclusion}

We presented \toolname, a novel relevance-feedback-driven technique and tool for visual pattern retrieval in multivariate time series.
Our system features an \acs{lsh}-based classifier with \textbf{massive scaling potential} that the analyst can \textbf{actively steer} and \textbf{easily understand}.
We found that the \acs{lsh}-based algorithm scales significantly better to large datasets than the state-of-the-art techniques, like \ac{dtw} and \ac{sax}, while retaining comparable accuracy compared to autoencoder-based systems, like PEAX, paving the path for real-time user interaction. 
Furthermore, users can understand our algorithm through the interactive query interface and effectively steer the modeling process.
Therefore, the prevailing problem of the subjectiveness of similarity cursing virtually all automatic similarity search algorithms can be well addressed. 
In the future, we can expect that our technical contributions, such as \toolname's adaptive classifier and visual similarity definition interfaces, will enable a better, more user-centric approach towards anomaly detection and motif discovery even in challenging data environments like multivariate or high-dimensional data spaces.



\clearpage


\bibliographystyle{abbrv-doi-hyperref-narrow}

\bibliography{pseudobib}

\end{document}


\onecolumn
\maketitle
\tableofcontents

\begin{acronym}
    \acro{dba}[DBA]{Dynamic Time Warping Barycenter Averaging}
    \acro{dtw}[DTW]{Dynamic Time Warping}
    \acro{dtwd}[DTW\textsubscript{D}]{Dynamic Time Warping Dependent}
    \acro{ed}[ED]{Euclidean Distance}
    \acro{lsh}[LSH]{Locality-Sensitive Hashing}
    \acro{mts}[MTS]{Multivariate Time Series}
    \acro{paa}[PAA]{Piecewise Aggregate Approximation}
    \acro{sax}[SAX]{Symbolic Aggregate approXimation}
    \acro{qalsh}[QALSH]{Query-Aware Locality Sensitive Hashing}
    \acro{uts}[UTS]{Univariate Time Series}
\end{acronym}
\clearpage

\appendix

\raggedbottom

\renewcommand\thefigure{\thesection.\arabic{figure}} 
\setcounter{figure}{0}

\clearpage
\section{Mathematical Proof for Model Constraint}
\label{sec:mathproof}

To avoid vanishing and exploding parameters, we want to retain the expectation of the magnitude of the hash function vector $ \vec{a} $, namely $ E(||\vec{a}||) $.
For convenience, we choose to constrain $ E(||\vec{a}||^2) $, namely we desire $ E(||\vec{b}||^2) = E(||\vec{a}||^2) $.

This is achieved through normalization of $ \vec{w} $.

\begin{equation} 
a_j \sim \mathcal{N}(0, 1)
\end{equation}

\begin{equation} 
b_j = w_j*a_j \sim \mathcal{N}(0, w_j^2) 
\end{equation}

\begin{equation} 
\label{eq:maga}
E(||\vec{a}||^2) 
    = E(\sum_{j=1}^d a_j^2) 
    = \sum_{j=1}^d E(a_j^2) 
    = \sum_{j=1}^d E((a_j - 0)^2)
    = \sum_{j=1}^d E((a_j - E(a_j))^2)
    = \sum_{j=1}^d Var(a_j)
    = \sum_{j=1}^d 1
    = d
\end{equation}


Following the same procedure as \autoref{eq:maga}, we get 

\begin{equation} 
E(||\vec{b}||^2) = \sum_{j=1}^d Var(b_j) = \sum_{j=1}^d w_j^2 = E(||\vec{w}||^2) 
\end{equation}

Because we require that 

\begin{equation} 
 E(||\vec{b}||^2) = E(||\vec{a}||^2) 
\end{equation}

and 
\begin{equation} 
E(||\vec{w}||^2) = E(||\vec{b}||^2) = E(||\vec{a}||^2) = d
\end{equation}

should hold. 
Hence, we normalize 

\begin{equation} 
||\vec{w}|| \textrm{  to  }  \sqrt{d}
\end{equation}

\clearpage
\section{Experiment Setup}
\label{sec:app_experiment_setup}

\subsection{Hardware and Software}

We conduct all experiments on the same laptop with 
\begin{itemize}
    \item CPU: Intel® Core™ i5-8350U @ 1.70GHz with 4 cores and 8 threads
    \item RAM: 16GB
    \item HDD: 512GB SSD
    \item OS: 64-bit Ubuntu 20.04.1 LTS
\end{itemize}
Important frameworks and libraries include
\begin{flushleft}
    \begin{itemize}
        \item \toolname (\url{https://git.science.uu.nl/vig/sublinear-algorithms-for-va/locality-sensitive-hashing-visual-analytics})
            \begin{itemize}
                \item Python
                \item C++
                \item Angular (\url{https://angular.io/})
                \item Flask (\url{https://flask.palletsprojects.com/en/1.1.x/})
                \item plotly (\url{https://plotly.com})
                \item ZingChart (\url{https://www.zingchart.com/})
            \end{itemize}
        \item \acf{dtw}: from ucrdtw (\_ucrdtw.ucrdtw, \url{https://github.com/klon/ucrdtw}), only works for univariate time series
        \item \acf{dtwd}: from tslearn (tslearn.metrics.dtw \url{https://github.com/tslearn-team/tslearn})
        \item \acf{dba}: from \url{https://github.com/fpetitjean/DBA}
        \item \acf{sax}: also from tslearn (tslearn.piecewise.SymbolicAggregateApproximation)
        \item PEAX: \url{https://github.com/novartis/peax}
    \end{itemize}
\end{flushleft}

\subsection{Datasets}


\begin{figure}[H]
    \centering
    \includegraphics[width=\linewidth]{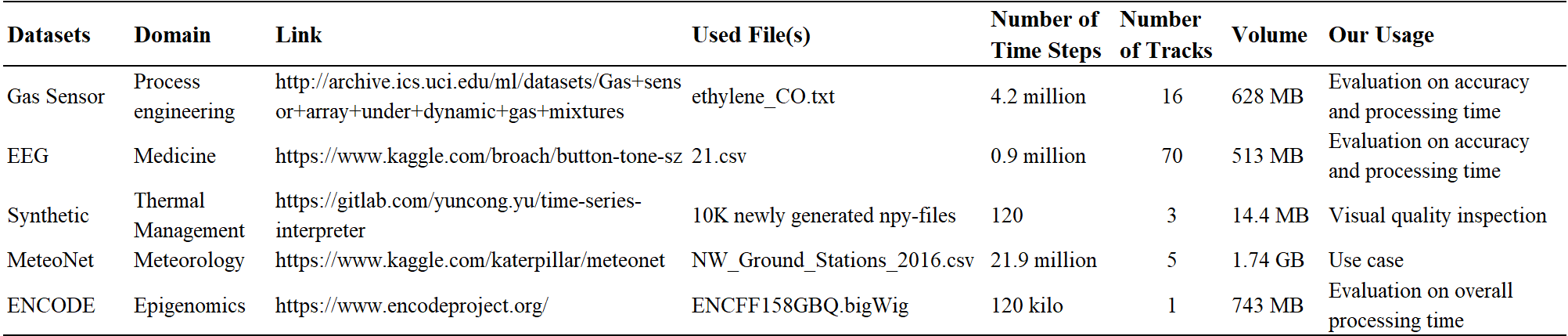}
    \caption{Datasets.}
    \label{fig:app_datasets}
\end{figure}

To draw a robust conclusion on the performance of our algorithm, we do our experiments using five different datasets:
\begin{itemize}
    \item \textbf{Epigenomic Dataset}\footnote{\url{https://www.encodeproject.org/files/ENCFF158GBQ/}}: This dataset contains time series data transcripted from gene data which is used to find functional elements of the human genome. We discretize the original dataset to 124,621 windows, with each window containing 120 timesteps and one track.
    \item \textbf{EEG Database}\footnote{\url{https://www.kaggle.com/broach/button-tone-sz}}: This dataset arises from a research studying abnormal predictive processes in schizophrenia. By using EEG the brain activity of dozen of patients is measured. For our experiments we took these measurements of one patient, which is a dataset containing 64 tracks with over 900,000 timesteps.
    \item \textbf{Gas Sensor Dataset}\footnote{\url{http://archive.ics.uci.edu/ml/datasets/Gas+sensor+array+under+dynamic+gas+mixtures}}: This dataset contains the acquired time series from 16 chemical sensors exposed to gas mixtures at varying concentration levels. Each measurement was constructed by the continuous acquisition of the 16-sensor array signals for a duration of about 12 hours without interruption, resulting in a time series length of roughly four million timesteps.
    \item \textbf{Synthetic Dataset}\footnote{\url{https://gitlab.com/yuncong.yu/time-series-interpreter}}: This dataset is synthetically generated. The time series generator creates an MTS by repeating three types of segments: linear, constant, and a step response of first-order systems, all of which are segments found frequently in thermal systems. By specifying input parameters, we can apply randomness such as noise and create an MTS of custom length and size. The dataset we generated is already segmented into a collection of 10,000 smaller time series of length 120 and three tracks.
    \item \textbf{Meteorological Dataset}\footnote{\url{https://www.kaggle.com/katerpillar/meteonet?select=NW_Ground_Stations}}: This dataset is contains meteorological data from multiple ground stations across France. We took the measurements of one of these ground stations over the course of one year, which contains 7 different tracks 88,197 timesteps.
\end{itemize}

The first dataset was chosen to compare our algorithm to PEAX, which is specialized for this type of data. The second and third datasets are of the same domain as the datasets used by Yu et al. in their evaluation of their LSH-based algorithm  insinuating that \textit{PSEUDo} should perform well for both types of data. We used the last two datasets for our visual evaluation: the visual analysis and case studies. The simplicity of the time series and the relatively small size of the datasets make the results easier to understand.

\subsection{Parameter Settings}

The parameters of the query length, number of samples, and number of tracks depend on the experiment performed. We mentioned the exact values of these parameters in the experiments section.

Some of the algorithms (including our own) require a choice of parameters. These choices are made according to the general rule of thumbs provided by each algorithm.

\begin{itemize}
    \item \textbf{DTW}: The Sakoe-Chiba bound is set to 5\% of the query length
    \item \textbf{PSEUDo}: The false negative rate $\delta$ is set to 0.05, approximation ratio $c$ to 1.3 and bucket size $\omega$ to 0.75$r$. For the DTW distance metric we apply the same Sakoe-Chiba bound as above. The learning rate $\alpha$ is set to 0.75.
    \item \textbf{SAX}: The amount of segments is equal to the query size and we choose a dictionary size of 10.
    \item \textbf{PEAX}: The data representation is created using PEAX's autoencoder trained on 120 kb window sizes with 1,000 bp binning\footnote{\url{https://zenodo.org/record/2609763#.YFy99P4o9H4}}.
\end{itemize}




\clearpage
\section{Technical Comparison Full Experiment Results}
\label{sec:app_experiment_result}

Technical comparison of the processing time, accuracy and steerability between \toolname, PEAX, \acfi{dtwd}, \acfi{ed} and \acfi{sax}. \ac{sax} is missing at some points because of memory issue.

\subsection{Experiment on Overall Processing Time }
\begin{figure}[H]
    \centering
    \includegraphics[width=0.6\linewidth,trim=15 15 10 5,clip]{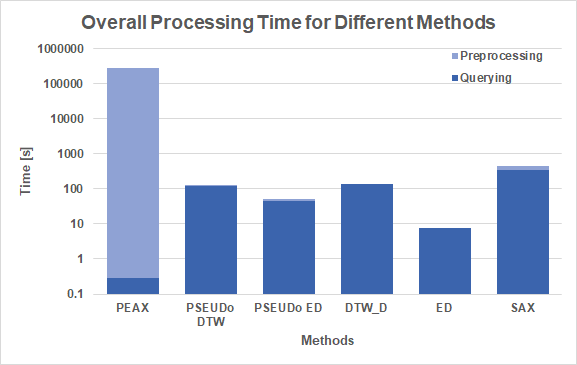}
    \caption{Overall processing time for different methods on the ENCODE dataset. }
\end{figure}

\clearpage
\subsection{Experiments on Processing Time}
\begin{figure}[H]
    \centering
    \includegraphics[width=0.55\linewidth,trim=1 1 1 1,clip]{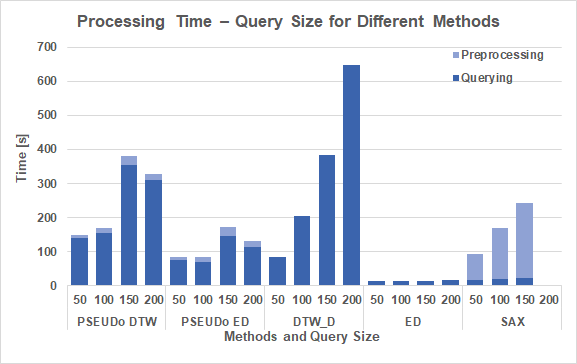}
    \caption{Processing time vs. query size for different methods on the gas sensor dataset. }
    \label{fig:app_speed_query_size}
\end{figure}

\begin{figure}[H]
    \centering
    \includegraphics[width=0.55\linewidth,trim=1 1 1 1,clip]{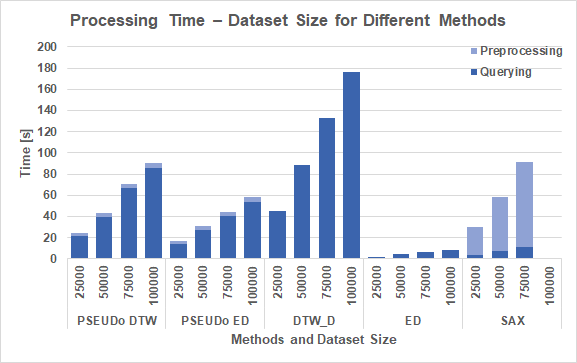}
    \caption{Processing time vs. dataset size for different methods on the EEG dataset. }
    \label{fig:app_speed_dataset_size}
\end{figure}

\begin{figure}[H]
    \centering
    \includegraphics[width=0.55\linewidth,trim=1 1 1 1,clip]{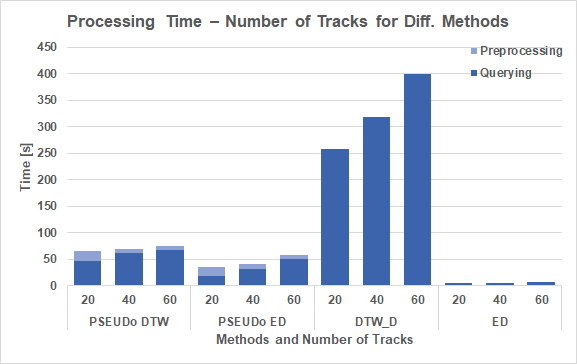}
    \caption{Processing time vs. number of tracks for different methods on the EEG dataset.}
    \label{fig:app_speed_number_of_tracks}
\end{figure}

\begin{landscape}
\begin{table}[]
\centering
\resizebox{\textwidth}{!}{%
\begin{tabular}{@{}lrrrrrrrrrrrrrrrrrrrr@{}}
\toprule
\multicolumn{21}{c}{\textbf{Speed Query Size (Seconds)}} \\ \midrule
 & \multicolumn{4}{c}{\textbf{PSEUDo DTW}} & \multicolumn{4}{c}{\textbf{PSEUDo ED}} & \multicolumn{4}{c}{\textbf{DTW}} & \multicolumn{4}{c}{\textbf{ED}} & \multicolumn{4}{c}{\textbf{SAX}} \\
 & \textbf{50} & \textbf{100} & \textbf{150} & \textbf{200} & \textbf{50} & \textbf{100} & \textbf{150} & \textbf{200} & \textbf{50} & \textbf{100} & \textbf{150} & \textbf{200} & \textbf{50} & \textbf{100} & \textbf{150} & \textbf{200} & \textbf{50} & \textbf{100} & \textbf{150} & \textbf{200} \\
\textbf{Preprocessing} & 9,8 & 15,1 & 26,1 & 16,4 & 9,8 & 15,1 & 26,1 & 16,4 & 0,0 & 0,0 & 0,0 & 0,0 & 0,0 & 0,0 & 0,0 & 0,0 & 77,5 & 149,6 & 220,6 & 0,0 \\
\textbf{Querying} & 139,6 & 153,6 & 355,6 & 310,3 & 76,0 & 70,0 & 145,5 & 114,9 & 84,9 & 204,6 & 384,2 & 648,4 & 12,8 & 13,5 & 14,9 & 15,8 & 17,2 & 19,5 & 22,6 & 0,0 \\
\textbf{Combined} & 149,4 & 168,7 & 381,7 & 326,7 & 85,9 & 85,2 & 171,5 & 131,3 & 84,9 & 204,6 & 384,2 & 648,4 & 12,8 & 13,5 & 14,9 & 15,8 & 94,7 & 169,1 & 243,2 &  \\ \bottomrule
\end{tabular}%
}
\end{table}

\begin{table}[]
\centering
\resizebox{\textwidth}{!}{%
\begin{tabular}{@{}lrrrrrrrrrrrrrrrrrrrr@{}}
\toprule
\multicolumn{21}{c}{\textbf{Speed Dataset Size (Seconds)}} \\ \midrule
 & \multicolumn{4}{c}{\textbf{PSEUDo DTW}} & \multicolumn{4}{c}{\textbf{PSEUDo ED}} & \multicolumn{4}{c}{\textbf{DTW}} & \multicolumn{4}{c}{\textbf{ED}} & \multicolumn{4}{c}{\textbf{SAX}} \\
 & \textbf{25000} & \textbf{50000} & \textbf{75000} & \textbf{100000} & \textbf{25000} & \textbf{50000} & \textbf{75000} & \textbf{100000} & \textbf{25000} & \textbf{50000} & \textbf{75000} & \textbf{100000} & \textbf{25000} & \textbf{50000} & \textbf{75000} & \textbf{100000} & \textbf{25000} & \textbf{50000} & \textbf{75000} & \textbf{100000} \\
\textbf{Preprocessing} & 3,1 & 3,7 & 4,2 & 5,2 & 3,1 & 3,7 & 4,2 & 5,2 & 0,0 & 0,0 & 0,0 & 0,0 & 0,0 & 0,0 & 0,0 & 0,0 & 25,8 & 50,9 & 80,5 & 0,0 \\
\textbf{Querying} & 21,2 & 39,8 & 66,7 & 85,6 & 13,6 & 26,8 & 40,2 & 53,5 & 44,7 & 88,1 & 132,9 & 176,1 & 2,1 & 4,3 & 6,3 & 8,6 & 3,9 & 7,5 & 11,3 & 0,0 \\
\textbf{Combined} & 24,3 & 43,5 & 70,9 & 90,8 & 16,8 & 30,5 & 44,3 & 58,8 & 44,7 & 88,1 & 132,9 & 176,1 & 2,1 & 4,3 & 6,3 & 8,6 & 29,7 & 58,4 & 91,7 &  \\ \bottomrule
\end{tabular}%
}
\end{table}

\begin{table}[]
\centering
\begin{tabular}{@{}lllllllllllll@{}}
\toprule
\multicolumn{13}{c}{\textbf{Speed Number Tracks (Seconds)}} \\ \midrule
 & \multicolumn{3}{c}{\textbf{PSEUDo DTW}} & \multicolumn{3}{c}{\textbf{PSEUDo ED}} & \multicolumn{3}{c}{\textbf{DTW}} & \multicolumn{3}{c}{\textbf{ED}} \\
 & \textbf{20} & \textbf{40} & \textbf{60} & \textbf{20} & \textbf{40} & \textbf{60} & \textbf{20} & \textbf{40} & \textbf{60} & \textbf{20} & \textbf{40} & \textbf{60} \\
\textbf{Preprocessing} & \multicolumn{1}{r}{17,1} & \multicolumn{1}{r}{7,8} & \multicolumn{1}{r}{7,4} & \multicolumn{1}{r}{17,1} & \multicolumn{1}{r}{7,8} & \multicolumn{1}{r}{7,4} & \multicolumn{1}{r}{0,0} & \multicolumn{1}{r}{0,0} & \multicolumn{1}{r}{0,0} & \multicolumn{1}{r}{0,0} & \multicolumn{1}{r}{0,0} & \multicolumn{1}{r}{0,0} \\
\textbf{Querying} & \multicolumn{1}{r}{47,6} & \multicolumn{1}{r}{61,1} & \multicolumn{1}{r}{67,3} & \multicolumn{1}{r}{17,5} & \multicolumn{1}{r}{32,6} & \multicolumn{1}{r}{50,0} & \multicolumn{1}{r}{258,4} & \multicolumn{1}{r}{318,3} & \multicolumn{1}{r}{399,6} & \multicolumn{1}{r}{4,6} & \multicolumn{1}{r}{6,2} & \multicolumn{1}{r}{7,6} \\
\textbf{Combined} & 64,7 & 68,9 & 74,6 & 34,6 & 40,4 & 57,3 & 258,4 & 318,3 & 399,6 & 4,6 & 6,2 & 7,6
\end{tabular}%
\end{table}
\end{landscape}

\clearpage
\subsection{Experiments on Accuracy}

We conducted our experiments with varying datasets and parameters. The precision and recall of \textit{PSEUDo} (and other metrics) are calculated in the following way:

\begin{itemize}
    \item \textit{Recall}: The amount of top-50 DTW windows present in the set classified as \textit{similar}
    \item \textit{Precision-50}: The amount of top-50 DTW windows present in the top-50 windows
    \item \textit{Precision-10\%}: The amount of top-50 DTW windows present in the top-10\% windows. We will refer to this as the approximate precision.
\end{itemize}

\begin{figure}[H]
    \centering
    \includegraphics[width=0.5\linewidth,trim=1 1 1 1,clip]{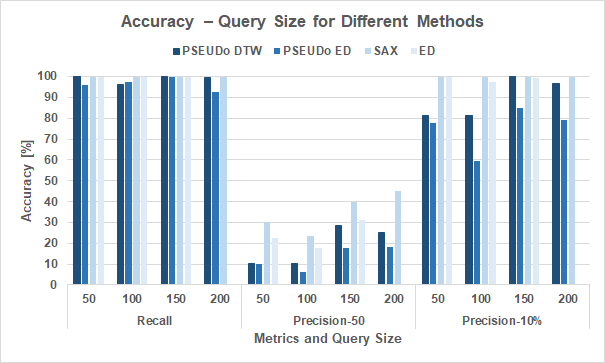}
    \caption{Accuracy vs. query size for different methods on the gas sensor dataset.}
    \label{fig:app_accuracy_query_size}
\end{figure}

\begin{figure}[H]
    \centering
    \includegraphics[width=0.5\linewidth,trim=1 1 1 1,clip]{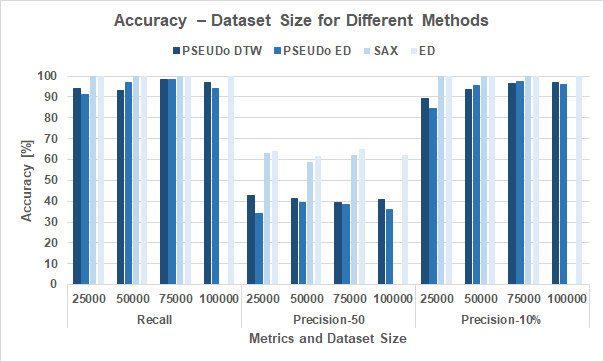}
    \caption{Accuracy vs. dataset size for different methods on the EEG dataset.}
    \label{fig:app_accuracy_dataset_size}
\end{figure}

\begin{figure}[H]
    \centering
    \includegraphics[width=0.5\linewidth,trim=1 1 1 1,clip]{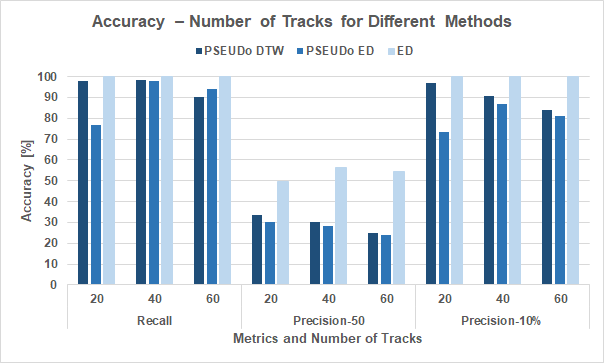}
    \caption{Accuracy vs. number of tracks for different methods on the EEG dataset.}
    \label{fig:app_accuracy_number_of_tracks}
\end{figure}

\begin{landscape}
\begin{table}[]
\centering
\resizebox{\textwidth}{!}{%
\begin{tabular}{@{}lrrrrrrrrrrrr@{}}
\toprule
\multicolumn{13}{c}{\textbf{Accuracy Query Size (in Percentage)}} \\ \midrule
 & \multicolumn{3}{l}{\textbf{Recall}} & \multicolumn{3}{l}{\textbf{}} & \multicolumn{3}{l}{\textbf{}} & \multicolumn{3}{l}{\textbf{}} \\
\textbf{} & \textbf{50} & \textbf{100} & \textbf{150} & \textbf{200} & \textbf{50} & \textbf{100} & \textbf{150} & \textbf{200} & \textbf{50} & \textbf{100} & \textbf{150} & \textbf{200} \\
\textbf{PSEUDo DTW} & 99,9 & 96,4 & 100 & 99,7 & 10,4 & 10,4 & 28,3 & 25 & 81,2 & 81,1 & 99,9 & 96,8 \\
\textbf{PSEUDo ED} & 95,8 & 97,5 & 100 & 92,4 & 9,8 & 6,1 & 17,7 & 18 & 77,8 & 59,4 & 85,1 & 79,3 \\
\textbf{SAX} & 100 & 100 & 100 & 0 & 22,7 & 17,5 & 31,1 & 0 & 99,7 & 97,6 & 99,4 & 0 \\
\textbf{ED} & 100 & 100 & 100 & 100 & 30,2 & 23,6 & 39,7 & 44,9 & 100 & 100 & 100 & 99,8
\end{tabular}%
}
\end{table}

\begin{table}[]
\centering
\resizebox{\textwidth}{!}{%
\begin{tabular}{@{}lrrrrrrrrrrrr@{}}
\toprule
\multicolumn{13}{c}{\textbf{Accuracy Dataset Size (in Percentage)}} \\ \midrule
 & \multicolumn{3}{l}{\textbf{Recall}} & \multicolumn{3}{l}{\textbf{}} & \multicolumn{3}{l}{\textbf{}} & \multicolumn{3}{l}{\textbf{}} \\
\textbf{} & \textbf{25000} & \textbf{50000} & \textbf{75000} & \textbf{100000} & \textbf{25000} & \textbf{50000} & \textbf{75000} & \textbf{100000} & \textbf{25000} & \textbf{50000} & \textbf{75000} & \textbf{100000} \\
\textbf{PSEUDo DTW} & 94,2 & 92,9 & 98,2 & 96,8 & 42,8 & 41,1 & 39,5 & 40,8 & 89,2 & 93,5 & 96,6 & 97,1 \\
\textbf{PSEUDo ED} & 91,1 & 96,8 & 98,2 & 94 & 33,9 & 39,3 & 38,6 & 36,2 & 84,3 & 95,3 & 97,5 & 96 \\
\textbf{SAX} & 100 & 100 & 100 & 0 & 63,1 & 58,6 & 62,4 & 0 & 100 & 100 & 100 & 0 \\
\textbf{ED} & 100 & 100 & 100 & 100 & 64,2 & 61,5 & 65 & 62,3 & 100 & 100 & 100 & 100
\end{tabular}%
}
\end{table}

\begin{table}[]
\centering
\begin{tabular}{@{}lrrrrrrrrr@{}}
\toprule
\multicolumn{10}{c}{\textbf{Accuracy Number Tracks (in Percentage)}} \\ \midrule
 & \multicolumn{3}{l}{\textbf{Recall}} & \multicolumn{3}{l}{\textbf{Precision-50}} & \multicolumn{3}{l}{\textbf{Precision-10\%}} \\
\textbf{} & \textbf{20} & \textbf{40} & \textbf{60} & \textbf{20} & \textbf{40} & \textbf{60} & \textbf{20} & \textbf{40} & \textbf{60} \\
\textbf{PSEUDo DTW} & 98 & 98,4 & 90,3 & 33,7 & 30,4 & 24,8 & 96,8 & 90,7 & 83,8 \\
\textbf{PSEUDo ED} & 76,7 & 97,9 & 94 & 30,3 & 28,2 & 23,9 & 73,2 & 86,7 & 81 \\
\textbf{ED} & 100 & 100 & 100 & 49,8 & 56,4 & 54,3 & 100 & 100 & 100
\end{tabular}%
\end{table}
\end{landscape}

\clearpage
\subsection{Experiments on Steerability}

\subsubsection{Experiment 1 }
\begin{figure}[H]
    \centering
    \begin{subfigure}[b]{0.24\linewidth}
        \includegraphics[width=\linewidth,trim=3 3 0 0,clip]{figures/sec_5_evaluation/5.1_steerability_prototype_query_1.png}
        \caption{Prototypes and variance bands.}
    \end{subfigure}
    \begin{subfigure}[b]{0.324\linewidth}
        \includegraphics[width=\linewidth,trim=3 3 3 5,clip]{figures/sec_5_evaluation/5.1_steerability_line_chart_query_1.png}
        \caption{Impact of feedback: distance to query}
    \end{subfigure}
    \caption{Steerability evaluation for query 1.}
    \label{fig:app_steerability_1}
\end{figure}

\subsubsection{Experiment 2 }
\begin{figure}[H]
    \centering
    \begin{subfigure}[b]{0.24\linewidth}
        \includegraphics[width=\linewidth,trim=3 3 0 0,clip]{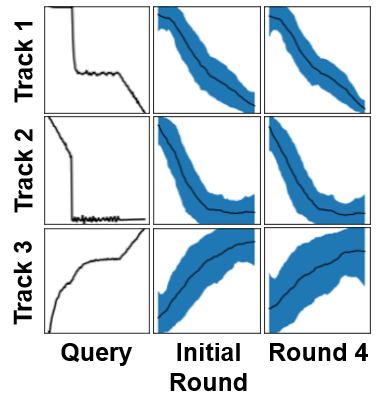}
        \caption{Prototypes and variance bands.}
    \end{subfigure}
    \begin{subfigure}[b]{0.324\linewidth}
        \includegraphics[width=\linewidth,trim=3 3 3 5,clip]{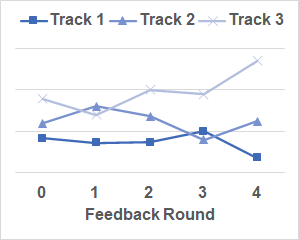}
        \caption{Impact of feedback: distance to query}
    \end{subfigure}
    \caption{Steerability evaluation for query 2}
    \label{fig:app_steerability_2}
\end{figure}

\subsubsection{Experiment 3 }
\begin{figure}[H]
    \centering
    \begin{subfigure}[b]{0.24\linewidth}
        \includegraphics[width=\linewidth,trim=3 3 0 0,clip]{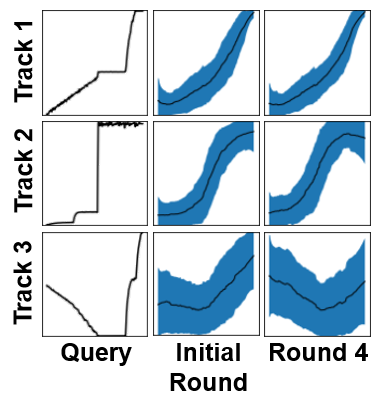}
        \caption{Prototypes and variance bands.}
    \end{subfigure}
    \begin{subfigure}[b]{0.324\linewidth}
        \includegraphics[width=\linewidth,trim=3 3 3 5,clip]{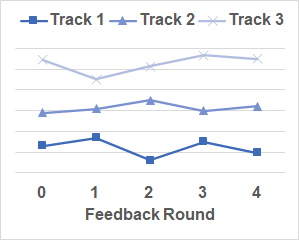}
        \caption{Impact of feedback: distance to query}
    \end{subfigure}
    \caption{Steerability evaluation for query 3}
    \label{fig:app_steerability_3}
\end{figure}

\subsubsection{Experiment 4 }
\begin{figure}[H]
    \centering
    \begin{subfigure}[b]{0.24\linewidth}
        \includegraphics[width=\linewidth,trim=3 3 0 0,clip]{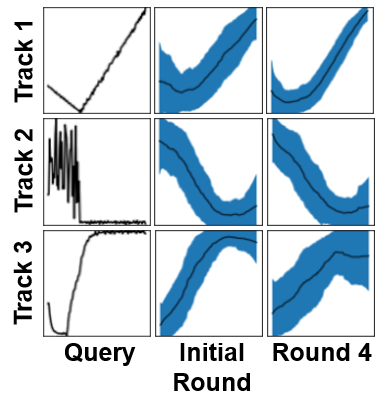}
        \caption{Prototypes and variance bands.}
    \end{subfigure}
    \begin{subfigure}[b]{0.324\linewidth}
        \includegraphics[width=\linewidth,trim=3 3 3 5,clip]{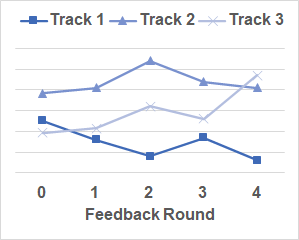}
        \caption{Impact of feedback: distance to query}
    \end{subfigure}
    \caption{Steerability evaluation for query 4}
    \label{fig:app_steerability_4}
\end{figure}

\subsubsection{Experiment 5 }
\begin{figure}[H]
    \centering
    \begin{subfigure}[b]{0.24\linewidth}
        \includegraphics[width=\linewidth,trim=3 3 0 0,clip]{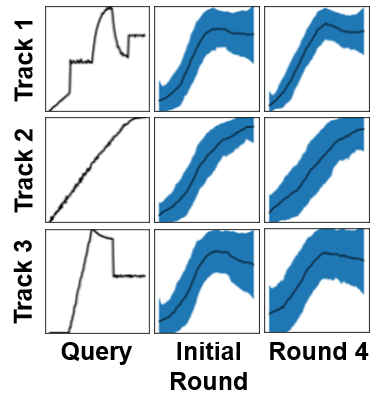}
        \caption{Prototypes and variance bands.}
    \end{subfigure}
    \begin{subfigure}[b]{0.324\linewidth}
        \includegraphics[width=\linewidth,trim=3 3 3 5,clip]{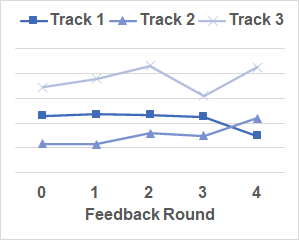}
        \caption{Impact of feedback: distance to query}
    \end{subfigure}
    \caption{Steerability evaluation for query 5.}
    \label{fig:app_steerability_5}
\end{figure}

\onecolumn
\clearpage
\section{Experiments on Visual Quality Inspection}

\subsection{Experiment 1}
\begin{figure}[H]
    \centering
    \includegraphics[width=\textwidth]{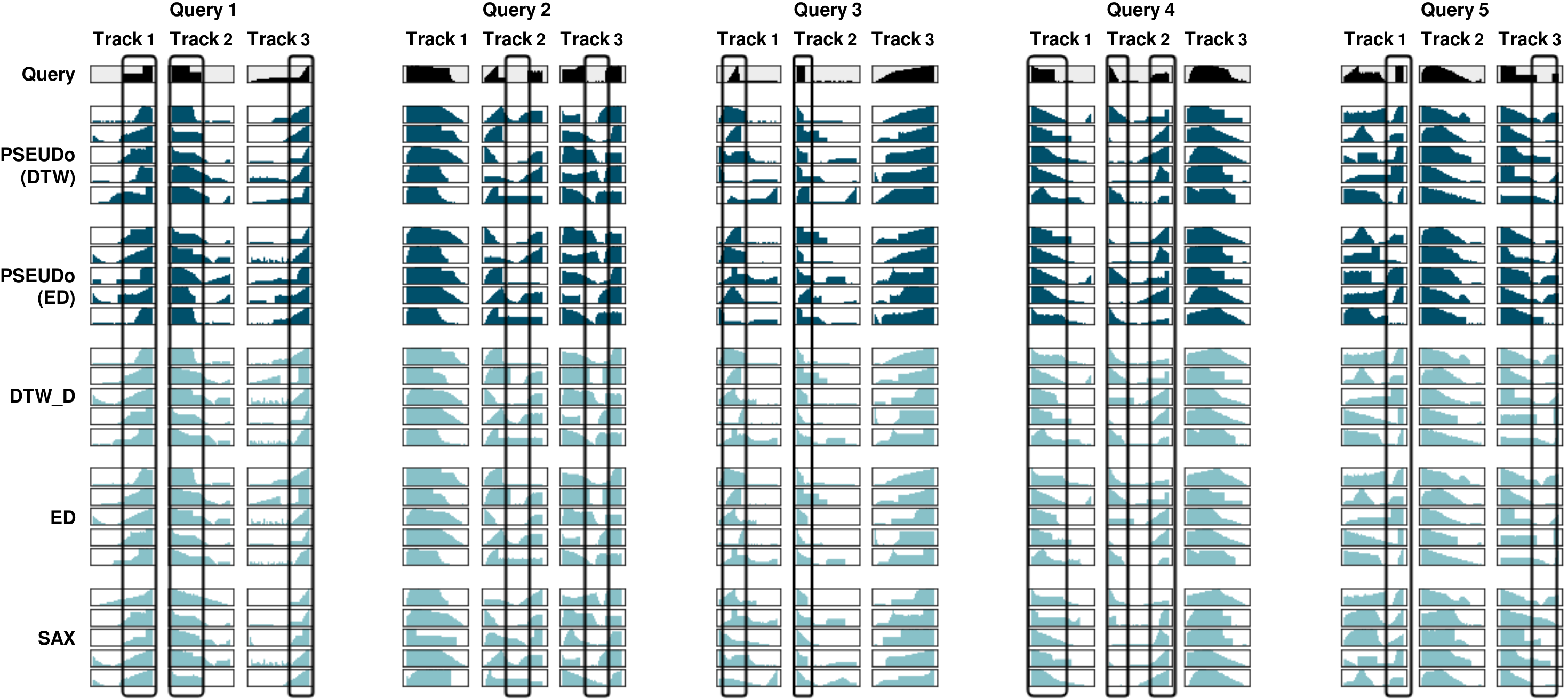}
    \caption{Visual comparison of 5-nearest neighbours for five 3-track queries between five techniques. Black boxes marks important features.}
\end{figure}

\subsection{Experiment 2}
\begin{figure}[H]
    \centering
    \includegraphics[width=\textwidth]{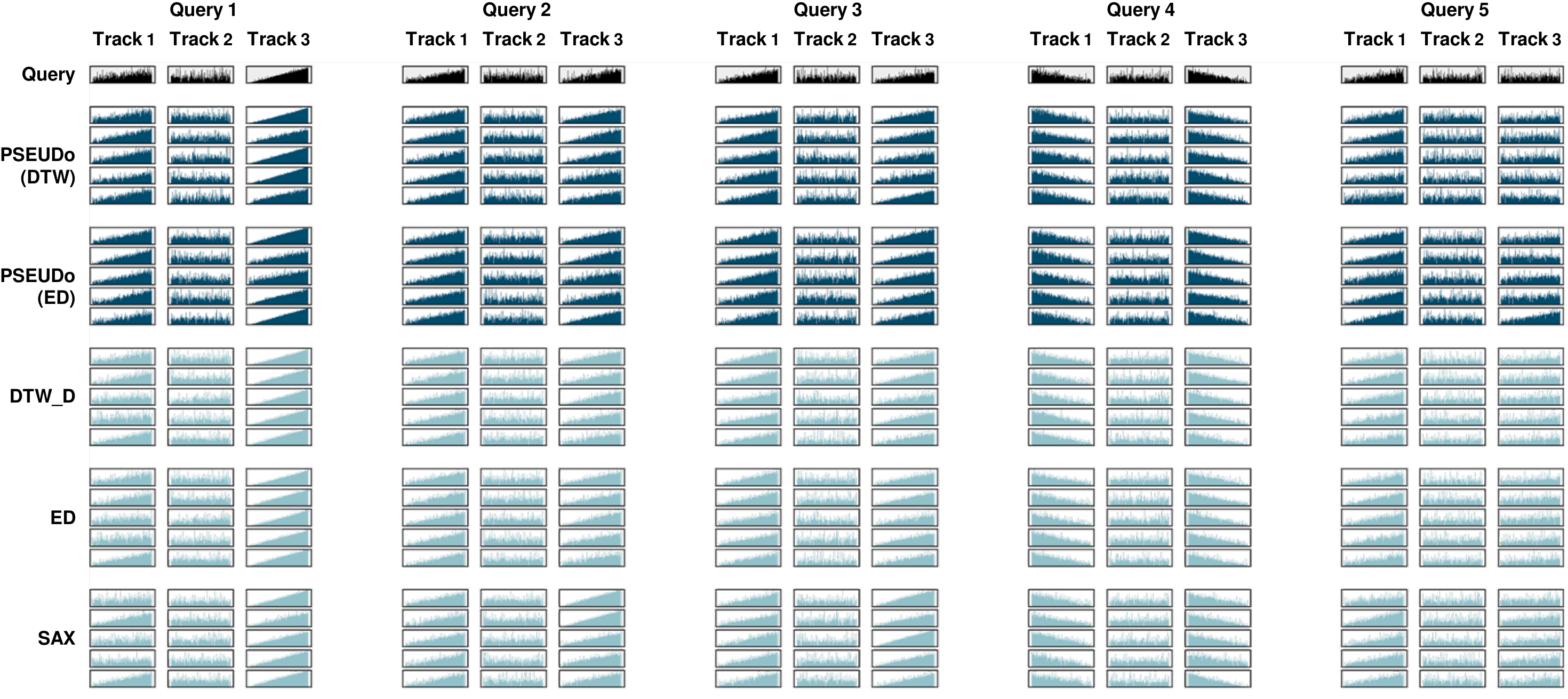}
    \caption{Visual comparison of 5-nearest neighbours for five 3-track queries between five techniques. Black boxes marks important features.}
\end{figure}

\clearpage
\section{Complete Use Cases}

We show one complete use case here.

\begin{figure}[h]
    \centering
    \includegraphics[width=\textwidth]{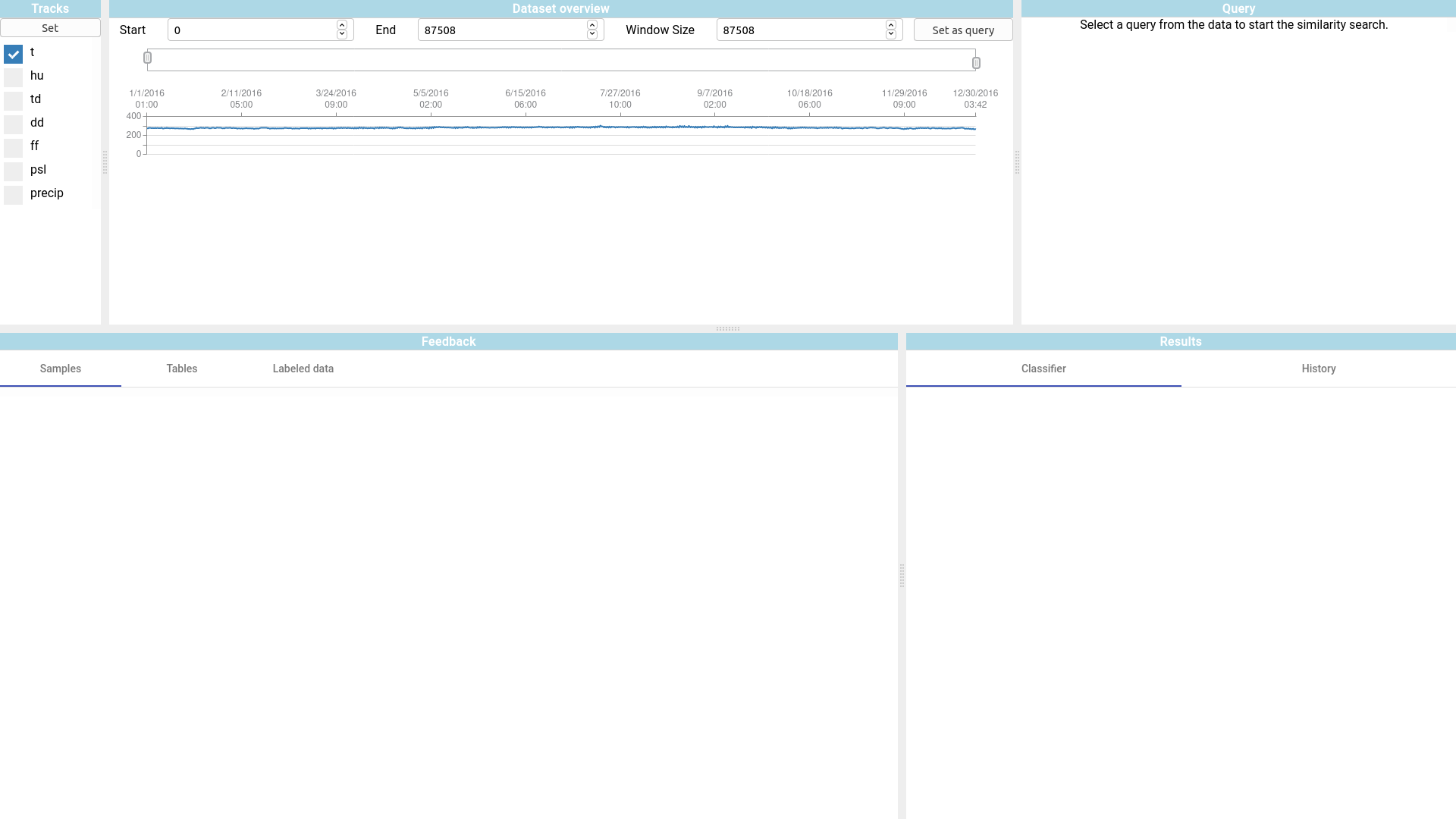}
    \caption{Jerry, a meteorologist wants to study cold wave from a meteorological dataset. He know it is related to a certain pattern in the data. He opens \toolname, load data. \toolname preselects the first track, which is temperature.}
\end{figure}

\begin{figure}[h]
    \centering
    \includegraphics[width=\textwidth]{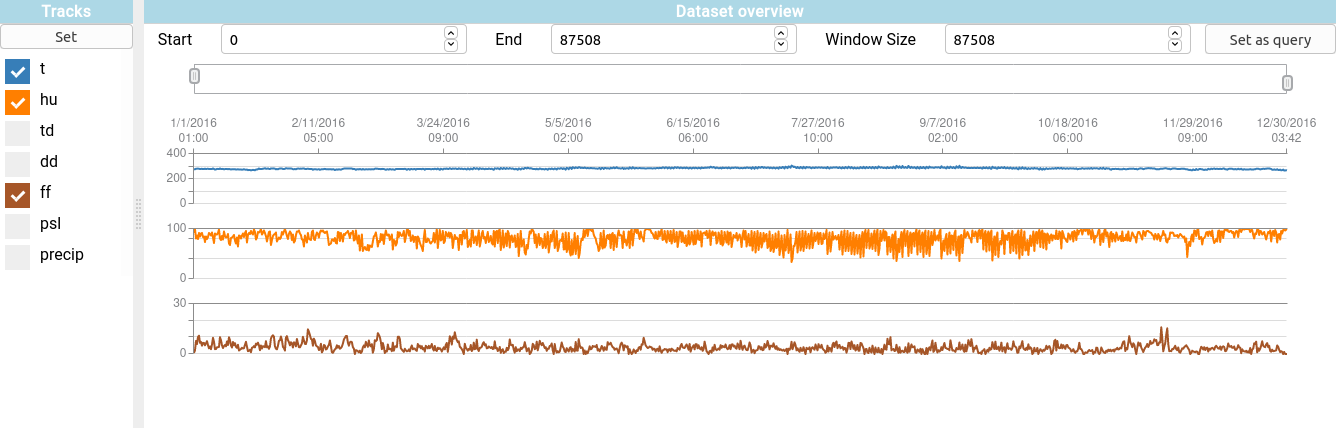}
    \caption{He selects other relevant tracks (humidity and wind speed) in the Tracks View and browse them in the Dataset Overview.}
\end{figure}

\begin{figure}[h]
    \centering
    \includegraphics[width=\textwidth]{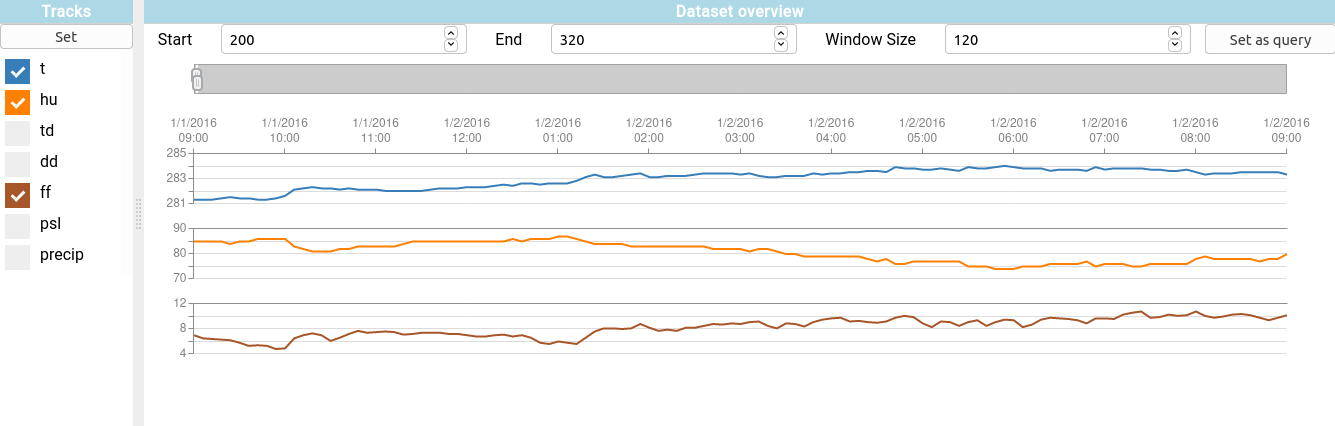}
    \caption{He finds an suspicious pattern and define it as the query by zooming in to it and clicks the ``Set as query" button.}
\end{figure}

\begin{figure}[h]
    \centering
    \includegraphics[width=\textwidth]{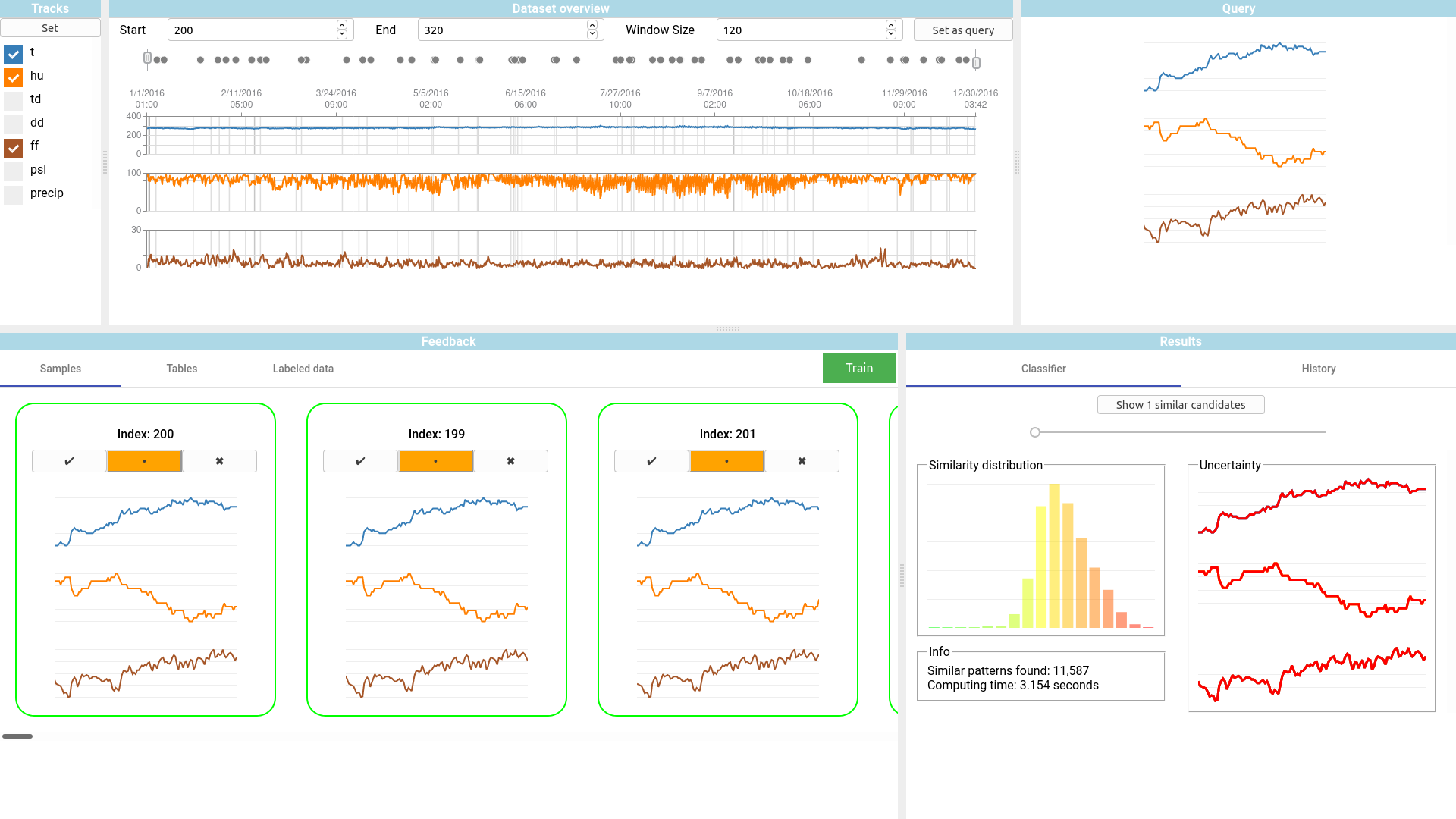}
    \caption{Query View immediately shows the current defined query. After several seconds, the Result View shows the similarity distribution of all time series windows in a histogram and the prototypes of all windows in each bin. The Samples View is filled with samples classified as similar or dissimilar by \ac{lsh}}
\end{figure}

\begin{figure}
    \centering
    \begin{subfigure}[b]{0.19\textwidth}
        \centering
        \includegraphics[width=\textwidth]{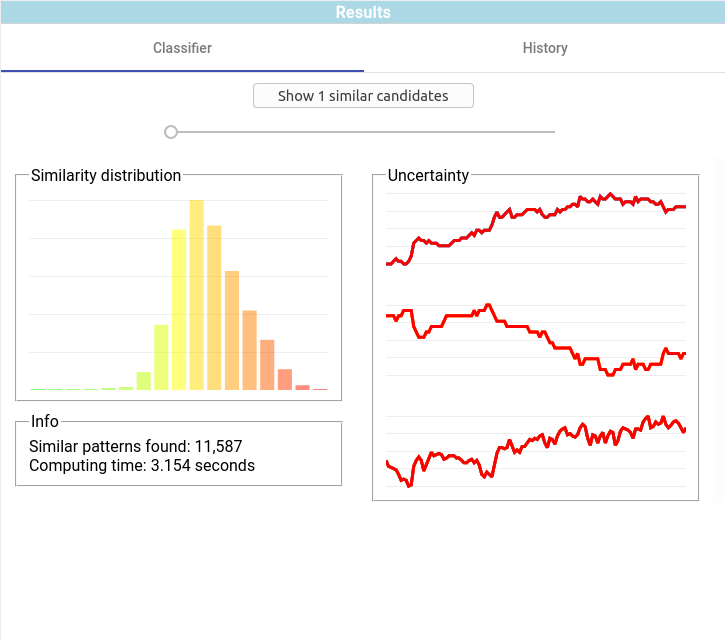}
        \caption{}
    \end{subfigure}
    \begin{subfigure}[b]{0.19\textwidth}  
        \centering 
        \includegraphics[width=\textwidth]{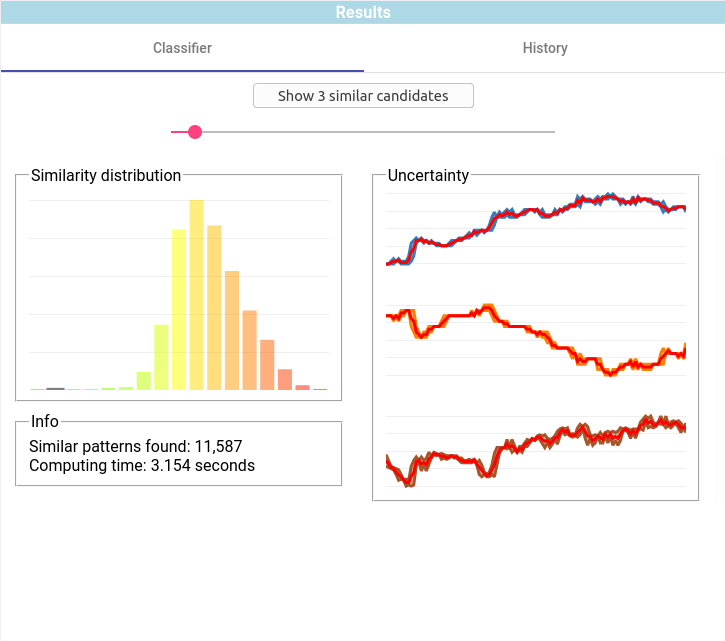}
        \caption{}
    \end{subfigure}
        \begin{subfigure}[b]{0.19\textwidth}  
        \centering 
        \includegraphics[width=\textwidth]{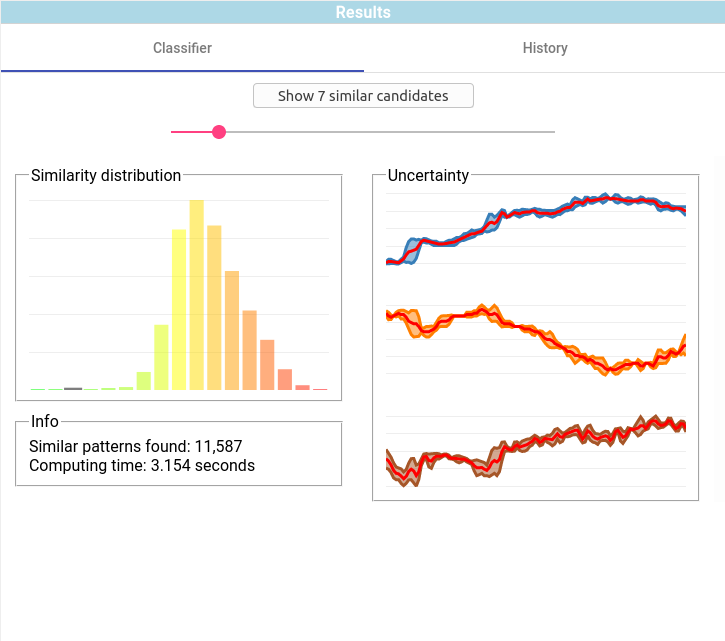}
        \caption{}
    \end{subfigure}
        \begin{subfigure}[b]{0.19\textwidth}  
        \centering 
        \includegraphics[width=\textwidth]{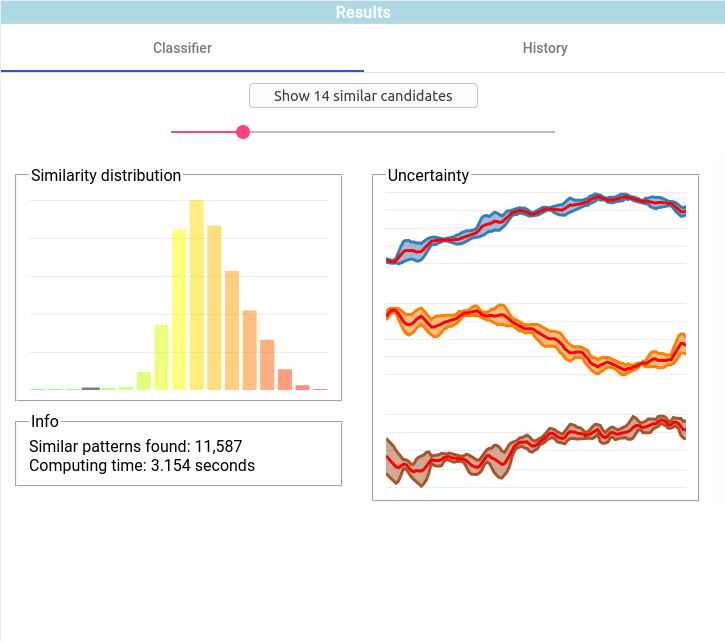}
        \caption{}
    \end{subfigure}
        \begin{subfigure}[b]{0.19\textwidth}  
        \centering 
        \includegraphics[width=\textwidth]{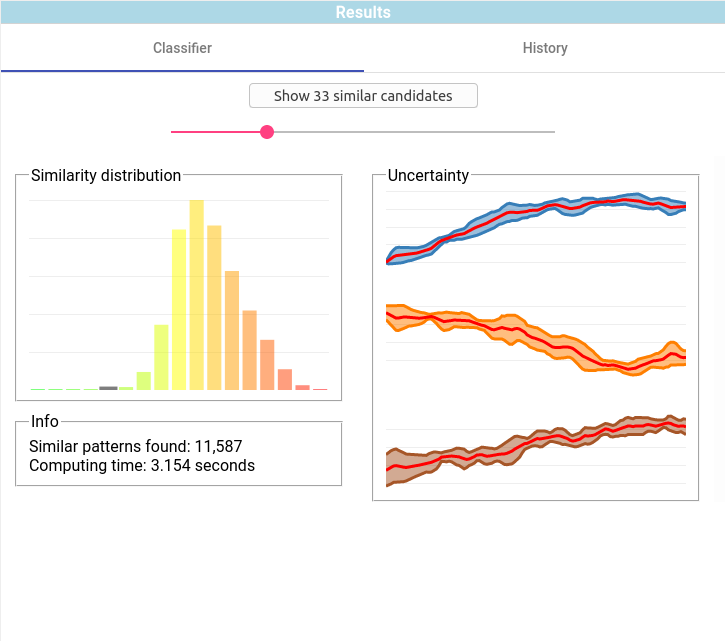}
        \caption{}
    \end{subfigure}
    \caption{He examines each bin in the histogram and finds that \toolname or the hash functions have overfitted the query.}
\end{figure}

\begin{figure}[h]
    \centering
    \includegraphics[width=\textwidth]{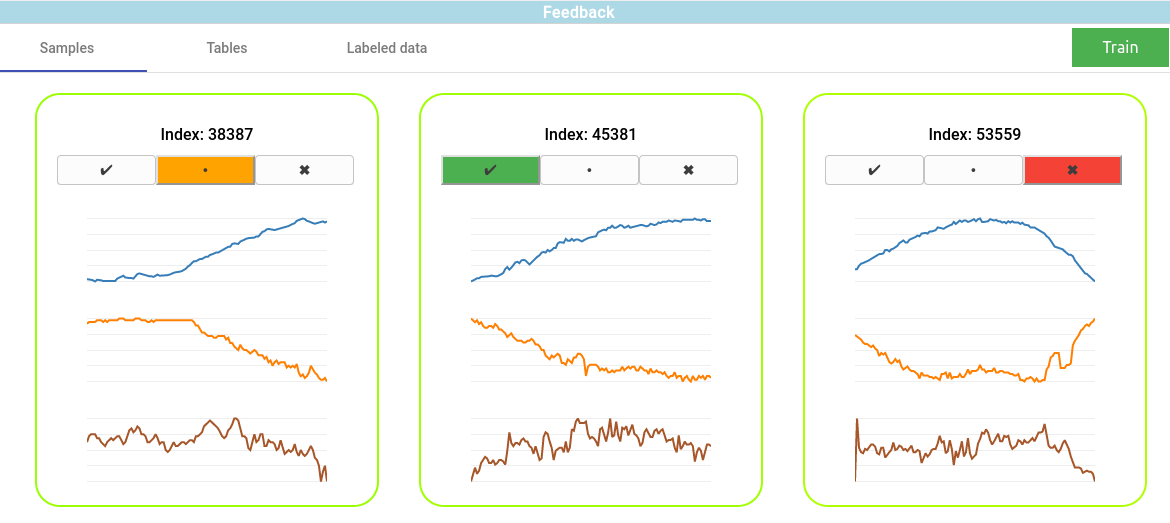}
    \caption{He examines the samples and rates them as similar, indecisive or not similar.}
\end{figure}

\begin{figure}[h]
    \centering
    \includegraphics[width=\textwidth, trim=0 0 10 0,clip]{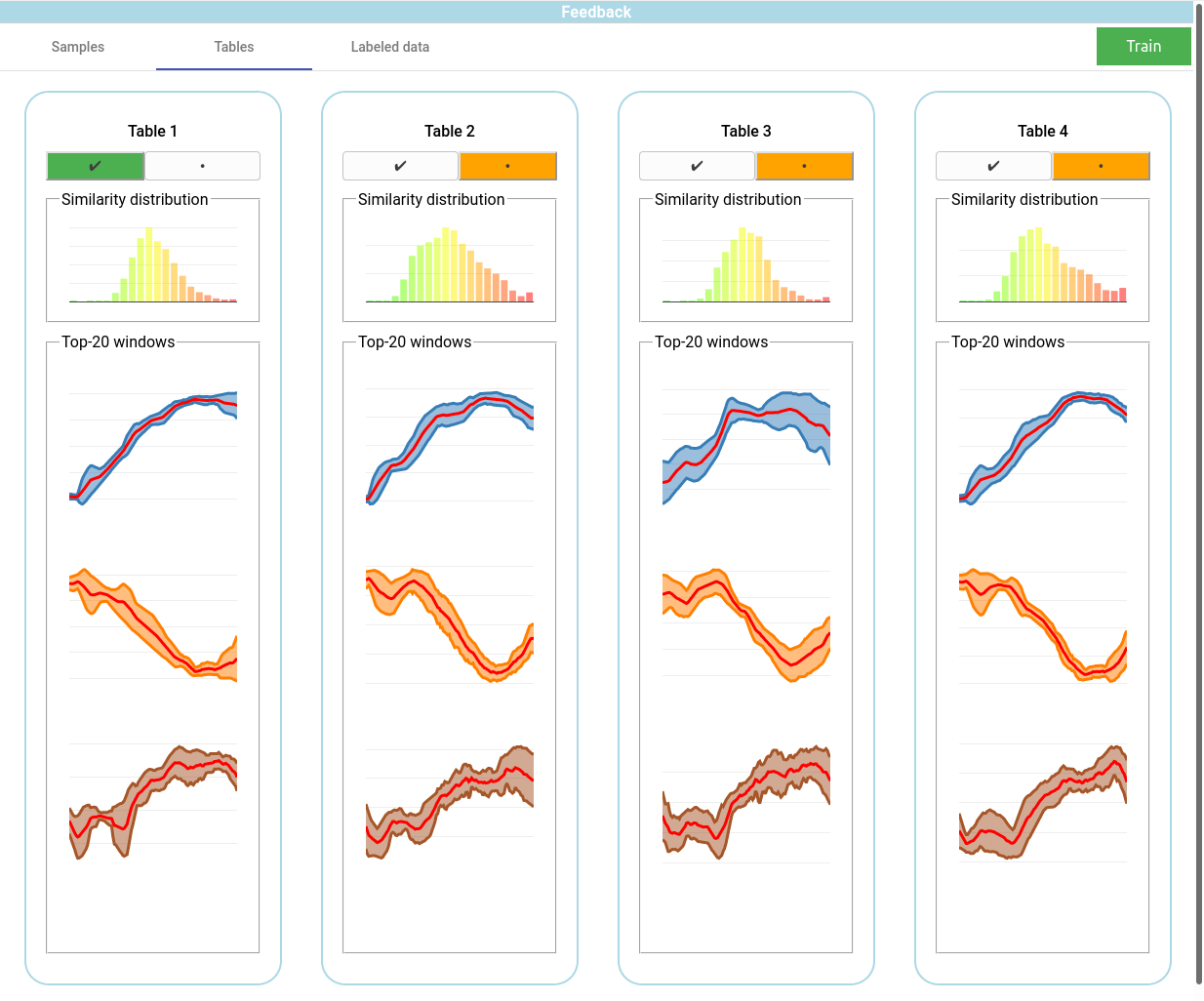}
    \caption{He moves on to check the classifiers / hash tables and similarly gives his binary feedback according to the similarity distribution and prototypes of the classifiers. Then, he clicks the ``Train" button.}
\end{figure}

\begin{figure}[h]
    \centering
    \includegraphics[width=\textwidth]{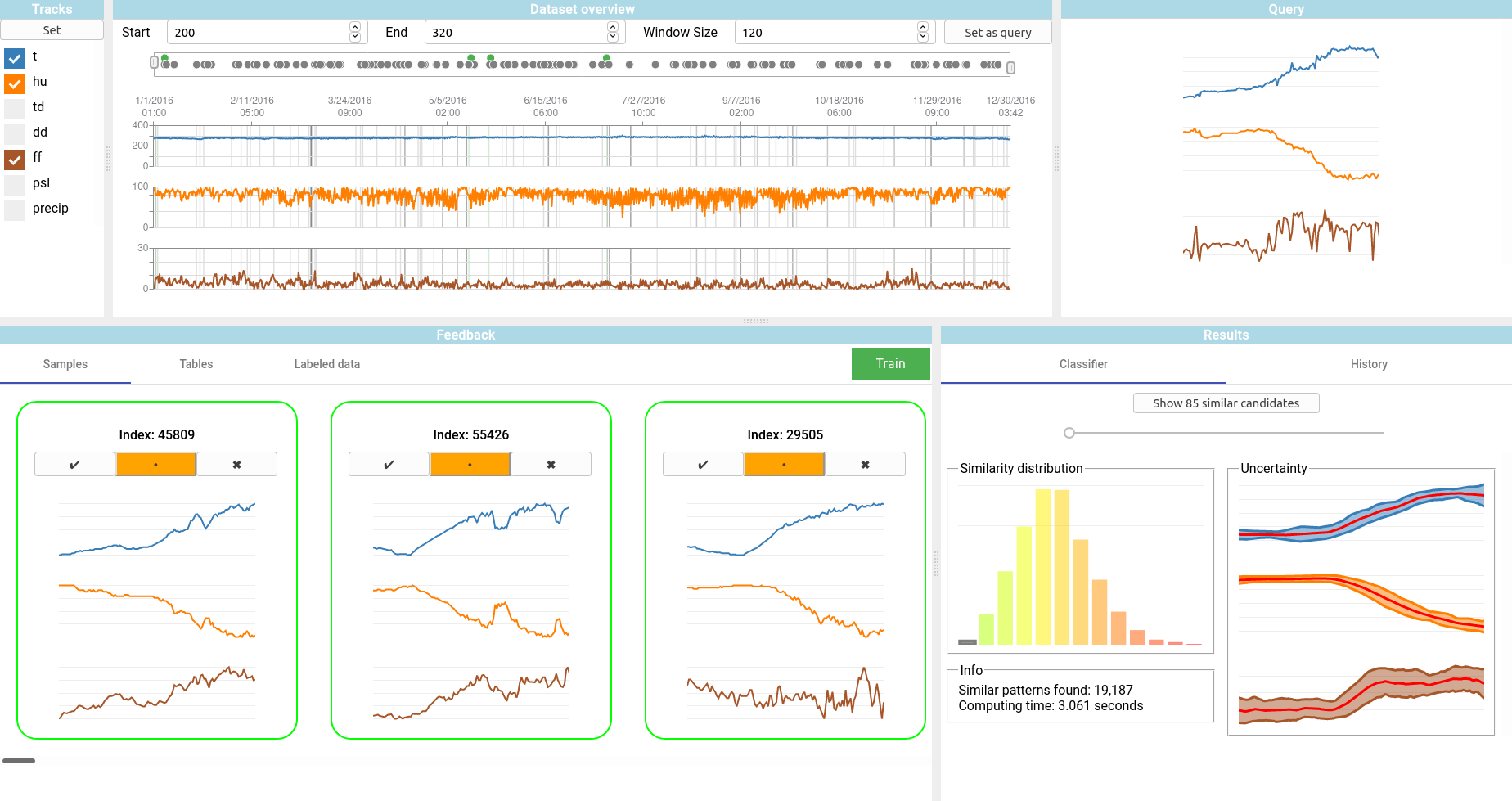}
    \caption{After few seconds, \toolname updates the results, as shown in the Result View and the Samples View.}
\end{figure}

\begin{figure*}
    \centering
    \begin{subfigure}[b]{0.19\textwidth}
        \centering
        \includegraphics[width=\textwidth]{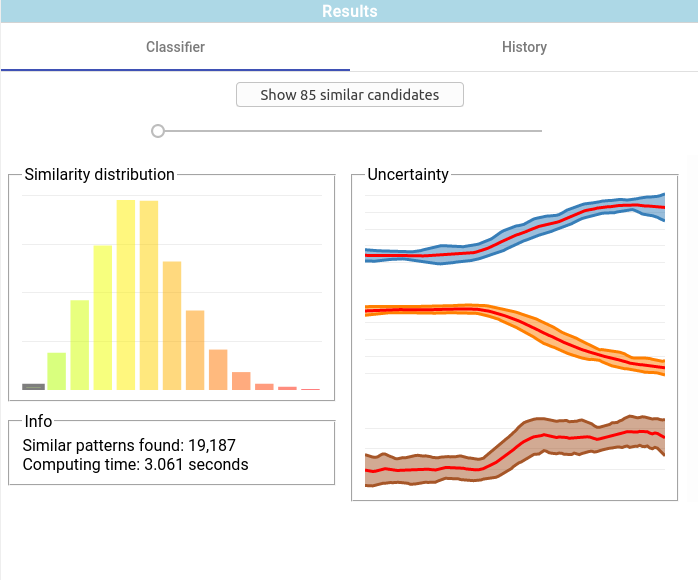}
        \caption{}
    \end{subfigure}
    \hfill
    \begin{subfigure}[b]{0.19\textwidth}  
        \centering 
        \includegraphics[width=\textwidth]{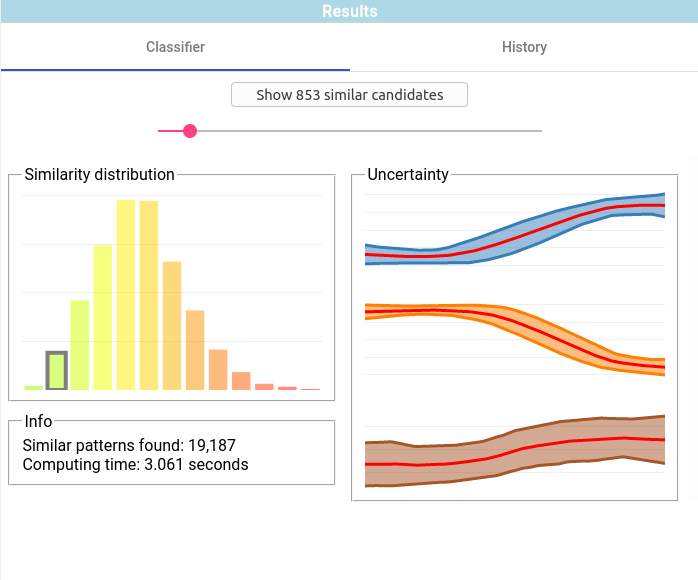}
        \caption{}
    \end{subfigure}
        \begin{subfigure}[b]{0.19\textwidth}  
        \centering 
        \includegraphics[width=\textwidth]{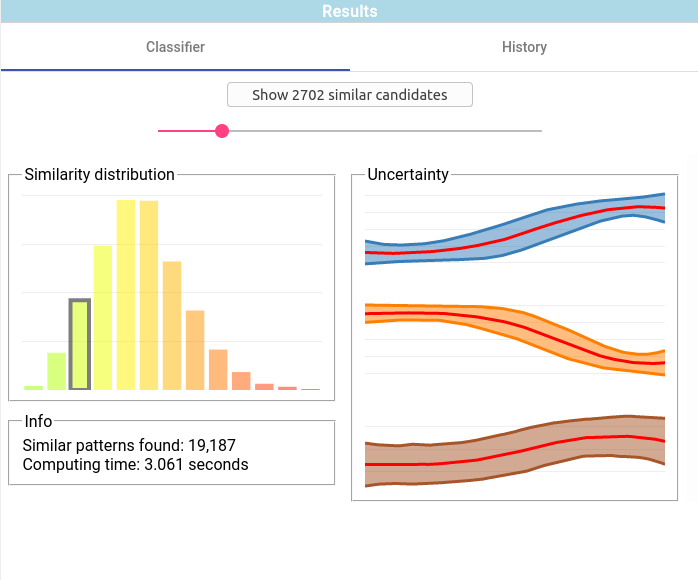}
        \caption{}
    \end{subfigure}
        \begin{subfigure}[b]{0.19\textwidth}  
        \centering 
        \includegraphics[width=\textwidth]{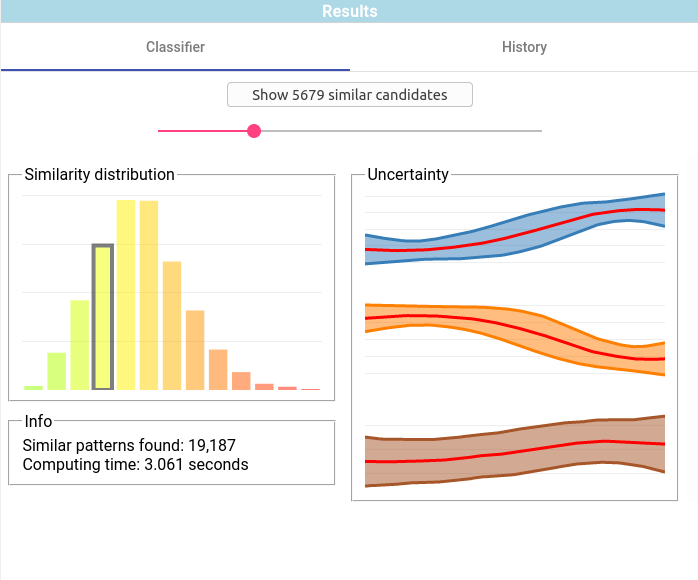}
        \caption{}
    \end{subfigure}
        \begin{subfigure}[b]{0.19\textwidth}  
        \centering 
        \includegraphics[width=\textwidth]{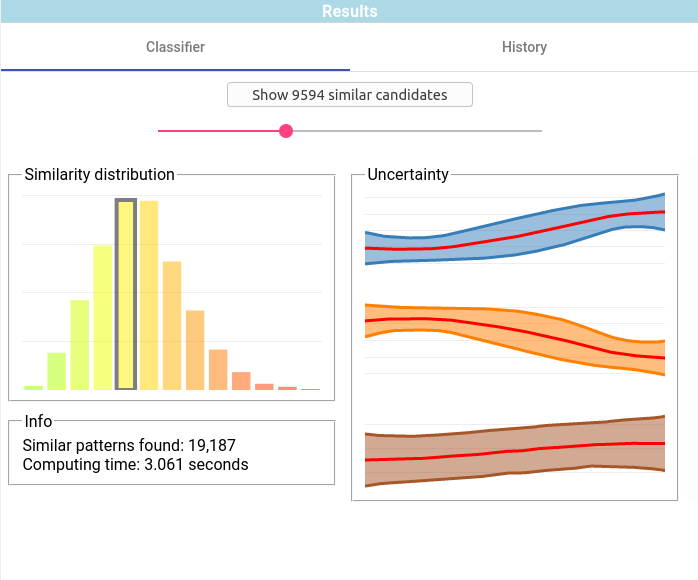}
        \caption{}
    \end{subfigure}
    \caption{He browsed the results, it goes in his desired direction, but there still seems to be room for improvement.}
\end{figure*}

\begin{figure}[h]
    \centering
    \includegraphics[width=\textwidth]{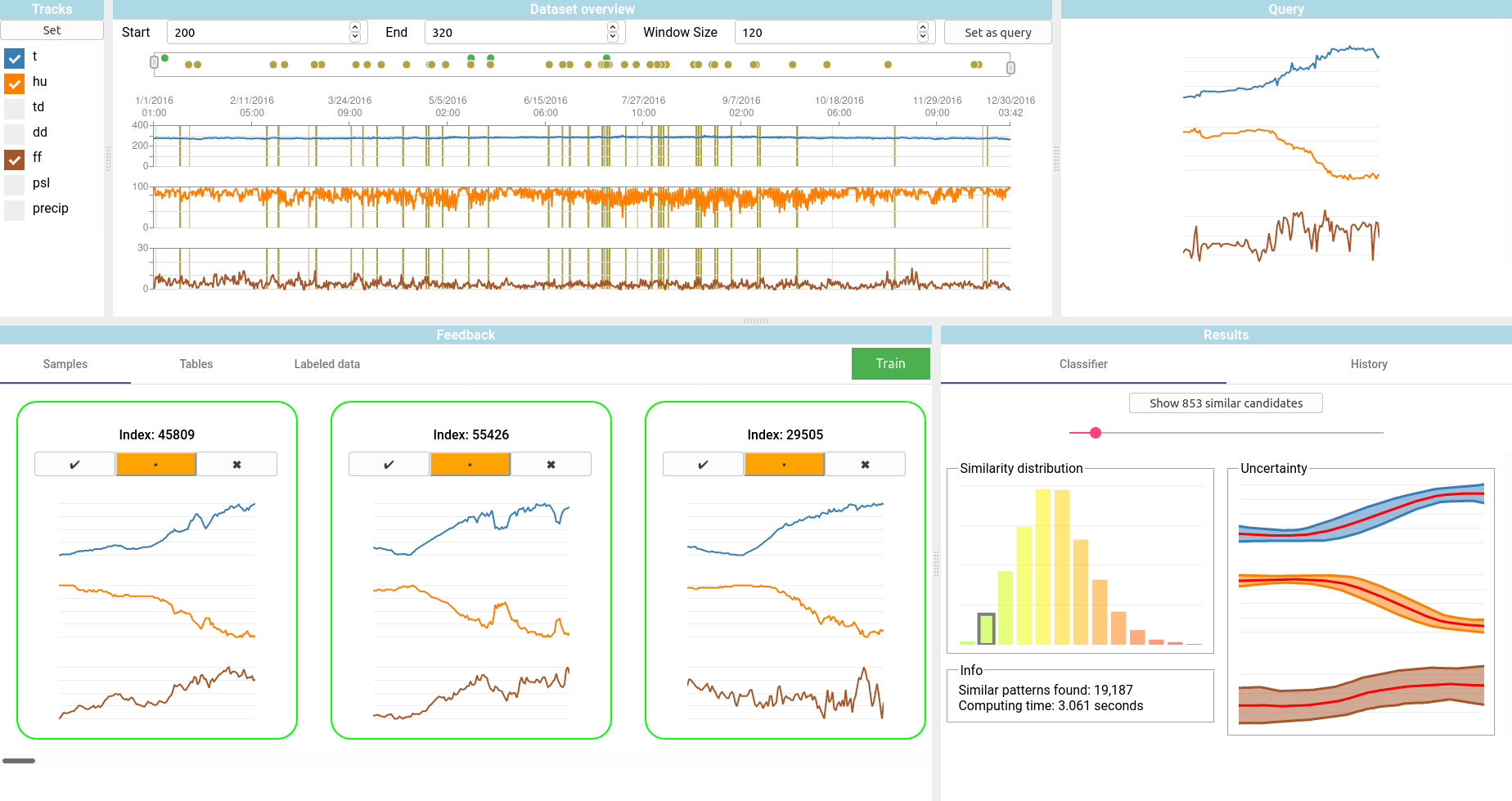}
    \caption{He continues several feedback rounds and is satisfied with the result. So, he selects a cut-off number of windows in the Result View and show all similar time series windows in the Dataset Overview.}
\end{figure}

\begin{figure}[h]
    \centering
    \includegraphics[width=\textwidth]{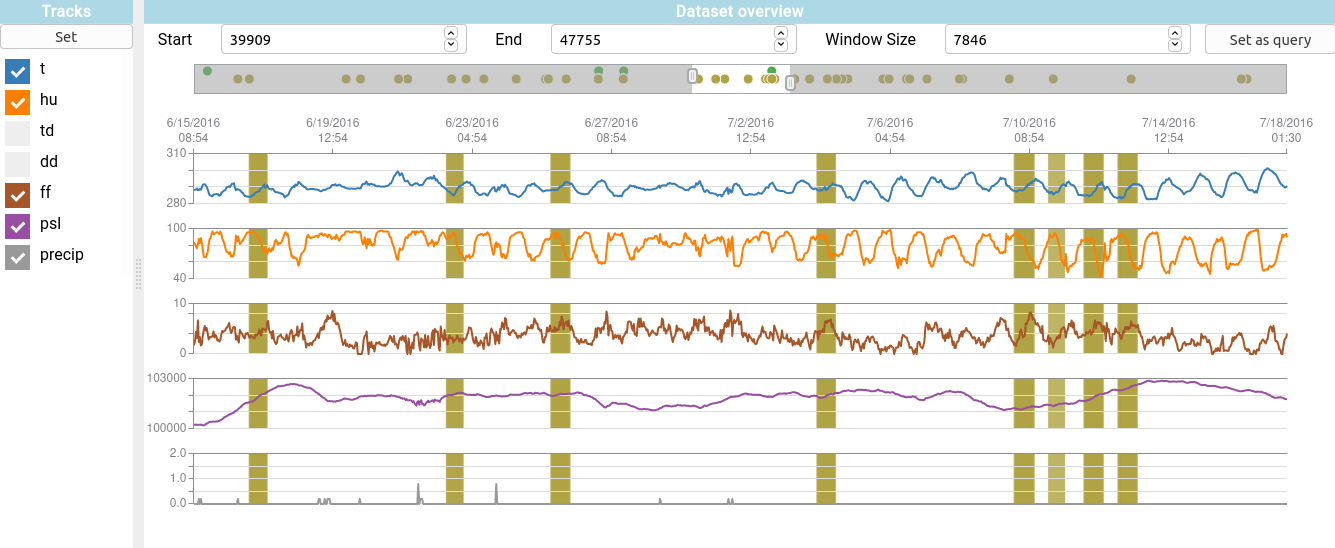}
    \caption{The found patterns are marked in the Dataset Overview. He goes on to examine them.}
\end{figure}